\documentclass[10pt,twocolumn,letterpaper,dvipsnames,table]{article}

\usepackage{cvpr}              

\definecolor{cvprblue}{rgb}{0.21,0.49,0.74}
\usepackage[pagebackref,breaklinks,colorlinks,allcolors=cvprblue]{hyperref}
\usepackage{makecell}
\usepackage{subcaption} 
\usepackage{pifont}
\usepackage{stmaryrd}
\usepackage{adjustbox} 
\usepackage[accsupp]{axessibility}
\def\paperID{8633} 
\def\confName{CVPR}
\def\confYear{2025}

\title{\raisebox{-1.0em}{\hspace{1em}\includegraphics[width=0.07\textwidth]{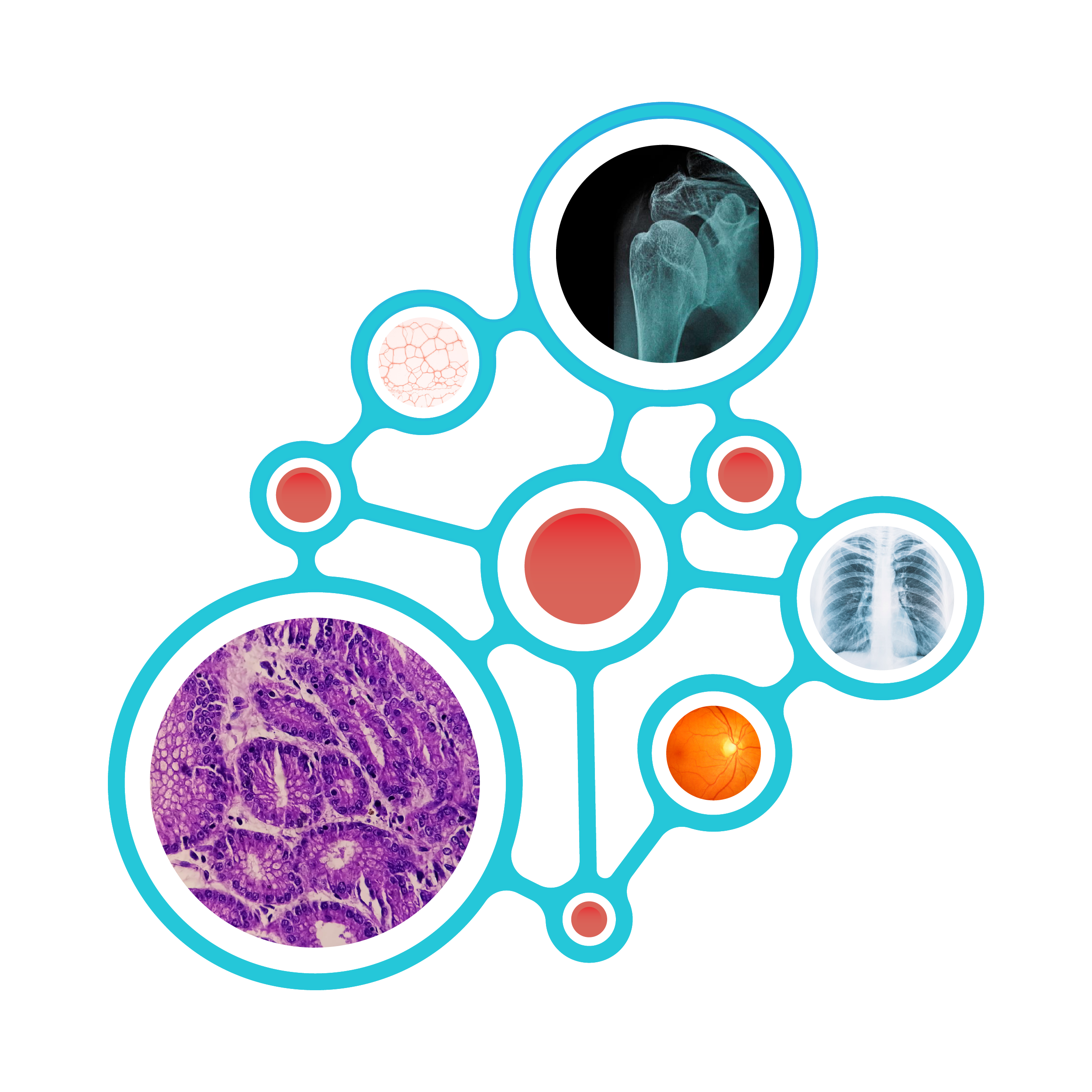}}
\dataset: An Open Biomedical Image-Caption Archive, Dataset, and  Vision-Language Models Derived from Scientific Literature }

\newcommand{\suppfigure}{%
    \renewcommand{\thefigure}{S\arabic{figure}}%
    \setcounter{figure}{0}%
}
\newcommand{\supptable}{%
    \renewcommand{\thetable}{S\arabic{table}}%
    \setcounter{table}{0}%
}

\author{Alejandro Lozano*\quad
Min Woo Sun*  \quad
James Burgess*\quad
Liangyu Chen  \quad
Jeffrey J. Nirschl\\
Jeffrey Gu\quad
Ivan Lopez\quad
Josiah Aklilu\quad
Anita Rau\quad
Austin Wolfgang Katzer\quad
Yuhui Zhang \\
Collin Chiu\quad
Xiaohan Wang \quad
Alfred Seunghoon Song \quad
Robert Tibshirani\quad 
Serena Yeung-Levy \\ \\
Stanford University
}

\newcommand{\dataset}{BIOMEDICA}

\newcommand{\numArticles}{6,042,494}

\newcommand{\numPairs}{24,076,288}
\newcommand{\numBenchmark}{40}

\newcommand{\rawDataSize}{27}
\newcommand{\numFigureReference}{30,711,542}

\newcolumntype{C}[1]{>{\centering\arraybackslash}p{#1}}

\definecolor{gen}{HTML}{FFFFFF}
\definecolor{med}{HTML}{FFFFFF}
\definecolor{bal}{HTML}{FFFFFF}

\newcommand{\modelGen}{BMC-CLIP}
\newcommand{\modelMed}{BMC-CLIP\textsubscript{CF}}
\newcommand{\modelBal}{BMC-CLIP\textsubscript{CB}}
\newcommand{\modelGenM}{\modelGen/WiSE-FT}
\newcommand{\modelMedM}{\modelMed/WiSE-FT}
\newcommand{\modelBalM}{\modelBal/WiSE-FT}

\begin{document}
\maketitle

\begin{abstract}
The development of vision-language models (VLMs) is driven by large-scale and diverse multi-modal datasets. However, progress toward generalist biomedical VLMs is limited by the lack of annotated, publicly accessible datasets across biology and medicine. Existing efforts are limited to narrow domains, missing  the full diversity of biomedical knowledge encoded in scientific literature. To address this gap, we introduce \dataset: a scalable, open-source framework to extract, annotate, and serialize the entirety of the PubMed Central Open Access subset into an easy-to-use, publicly accessible dataset. Our framework produces a comprehensive archive with over 24 million unique image-text pairs from over 6 million articles. Metadata and expert-guided annotations are additionally provided. \\ \\
We demonstrate the utility and accessibility of our resource by releasing \modelGen, a suite of CLIP-style models continuously pre-trained on \dataset~dataset via streaming (eliminating the need to download \rawDataSize\ TB of data locally). On average, our models achieve state-of-the-art performance across 40 tasks — spanning pathology, radiology, ophthalmology, dermatology, surgery, molecular biology, parasitology, and cell biology — excelling in zero-shot classification with 6.56\% average improvement (as high as 29.8\% and 17.5\% in dermatology and ophthalmology, respectively) and stronger image-text retrieval while using 10x less compute.
To foster reproducibility and collaboration, we release our codebase\footnote{\scriptsize{\url{https://github.com/minwoosun/biomedica-etl}}},\footnote{\scriptsize{\url{https://github.com/Ale9806/open_clip_with_biomedica}}} and dataset\footnote{\scriptsize{\url{https://huggingface.co/BIOMEDICA}} \\ $^*$ Equal contributions: \{lozanoe, minwoos, jmhb\}@stanford.edu} to the broader research community.

\end{abstract}


\begin{figure}[ht]
  \centering
  \vspace{-7mm}
  \includegraphics[width=\linewidth]{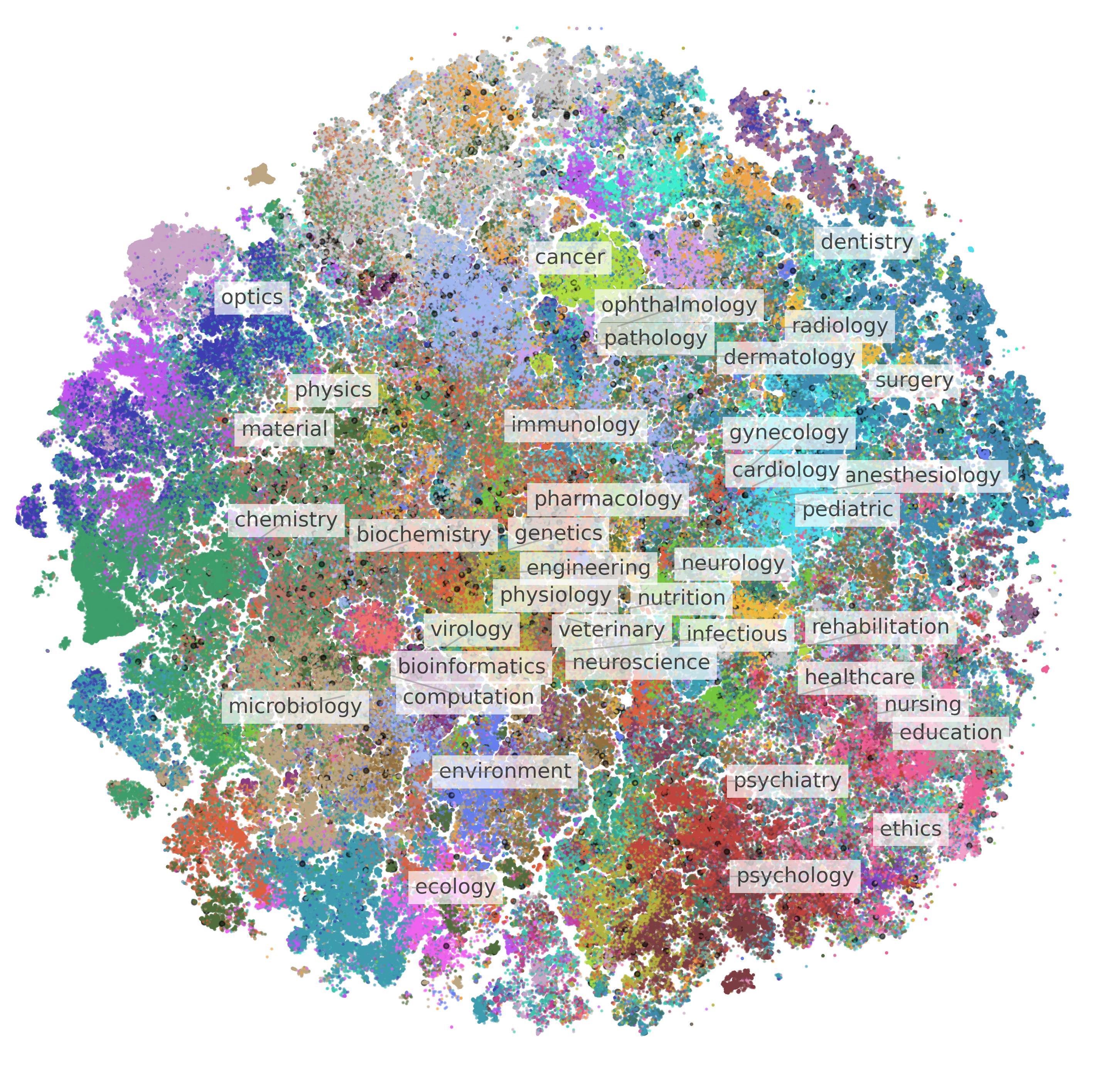}
  \vspace{-9.5mm}
  \caption{ Overlap of BIOMEDICA dataset with  the Landscape of Biomedical Research  \cite{gonzalez2024landscape} Each color region reflects thematic concentrations, capturing the diversity of topics within our dataset. Gray points represent articles not present in BIOMEDICA. }
  \label{fig:datamap}
\end{figure}

\section{Introduction}
\label{sec:intro}

Progress in vision-language foundation models (FMs) is driven by readily available large-scale and diverse datasets \cite{schuhmann2022laion,desai2021redcaps}. These datasets provide the basis for pretaining and adaptation,  leading to strong visual representations and expert-level zero-shot performance across a wide range of downstream tasks \cite{achiam2023gpt,touvron2023llama,yang2024qwen2}.

\begin{figure*}[h]
  \centering
  \includegraphics[width=0.9\textwidth]{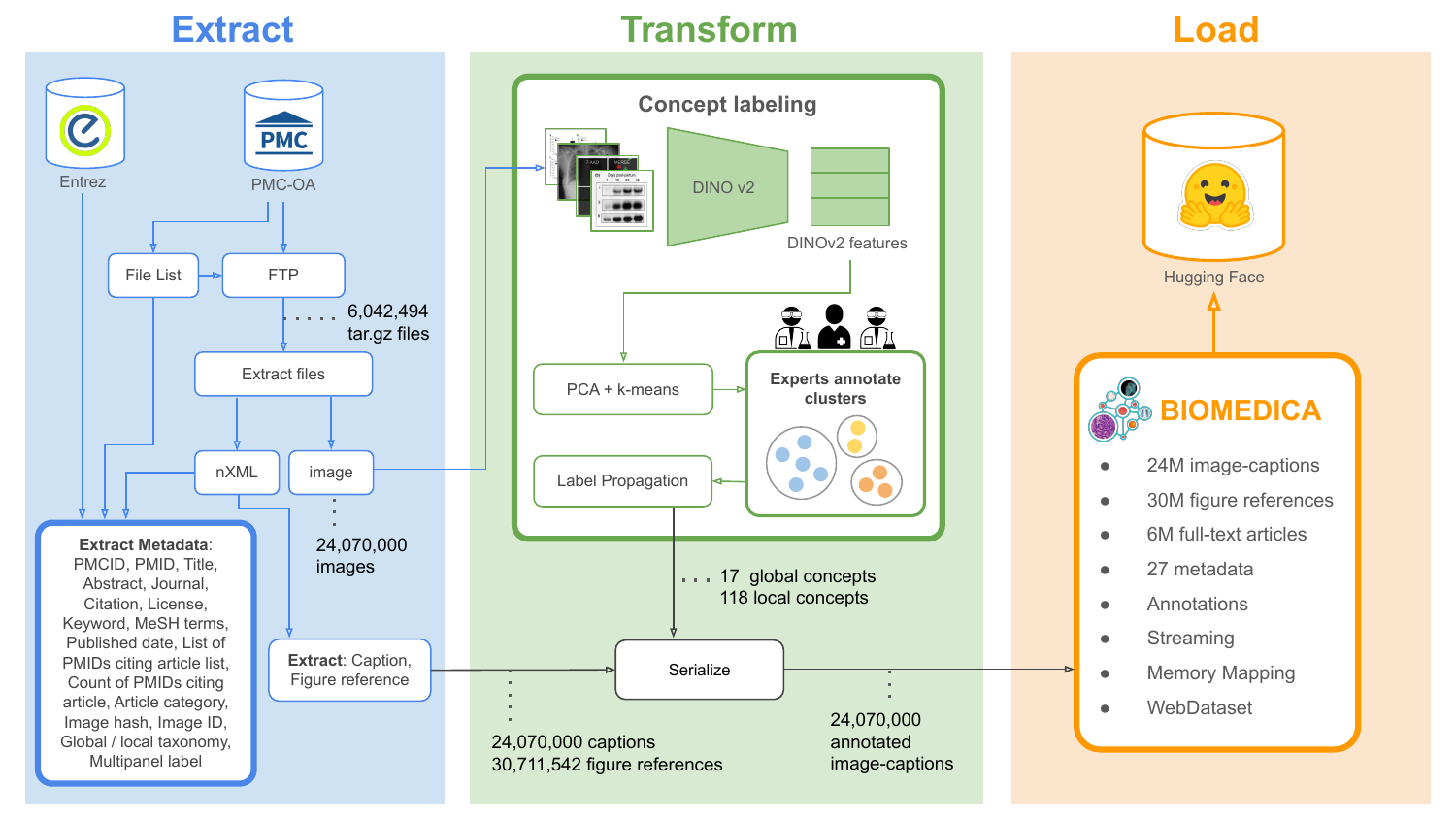}
  \vspace{-2.5mm}
  \caption{\dataset\ curation pipeline: In the Extract phase, metadata, text (caption, figure reference, full-text), and images are sourced and processed from PMC-OA. In the Transform phase, DINO v2 features are generated for each image, followed by clustering using PCA and k-means. Clinicians and scientists annotate these clusters, identifying 12 global concepts and 170 local concepts, which are then propagated across all images. Finally, in the Load phase, the dataset is made available on Hugging Face with the listed features.}
  \label{fig:data_workflow}
\end{figure*}

This success  has sparked interest in  biomedical multimodal FMs as integrating knowledge captured throughout different modalities across medical specialties, molecular biology, genetics, and related fields has the potential to revolutionize precision health. For instance, in daily practice, clinicians typically have one care-related question for every two patients seen \cite{del2014clinical}, often turning to sources like UpToDate for summarized information to address these inquiries \cite{daei2020clinical}. Questions that cannot be answered within three minutes are often abandoned,  negatively impacting patient care  \cite{del2014clinical,lozano2023clinfo}.  Biomedical FMs could bridge this gap by retrieving relevant information from visual and textual findings -- linking observations to emerging therapeutics, providing additional recommendations \cite{burgess2025microvqa} (e.g. linking a patient to a promising clinical trial \cite{jin2024matching,wornow2024zero}), flagging missed diagnoses \cite{saab2024capabilities}, and identifying biomarkers correlated with patient prognosis \cite{huo2024time} -- ultimately offering essential support for patient care and informed medical decision-making.

\begin{table*}[t]
\small
\centering
\begin{adjustbox}{max width=\textwidth}
\begin{tabular}{lcccccc}
\toprule
\rowcolor{gray!10}\textbf{Dataset}& \textbf{Scientific Papers} & \textbf{Image-Captions} & \textbf{Metadata Fields} & \textbf{Source} & \textbf{Streaming} & \textbf{Expert Curated Concepts} \\
ROCO \cite{pelka2018radiology}  & 1.8M & 0.08M & 3 & PMC-OA & \textbf{\texttimes} & \textbf{\texttimes} \\
MEDICAT \cite{subramanian2020medicat} & 1.8M & 0.21M & 3 & PMC-OA & \textbf{\texttimes} & \textbf{\texttimes} \\
PMC-OA \cite{lin2023pmc} & 2.4M & 1.65M & 0 & PMC-OA & \checkmark & \textbf{\texttimes} \\
PMC-15M \cite{zhang2023biomedclip} & 3M & 15M & 3 & PMC-OA & \textbf{\texttimes} & \textbf{\texttimes} \\
\midrule
\dataset{} (\textbf{Ours}) & 6M & 24M & 27$^{*}$ & PMC-OA & \checkmark & Clinicians \& Scientists \\
\bottomrule
\end{tabular}
\end{adjustbox}
\caption{Comparison of our work to existing literature-based biomedical datasets. BIOMEDICA stands out by offering 2x more scientific papers, 1.6x more image-caption pairs, and 9x more metadata (including expert-curated concepts). See Table \ref{tab:metadata-list} for a full list of  metadata.
 }
\label{xc}
\end{table*}


However, despite the potential and growing interest in generalist biomedical FMs \cite{moor2023foundation}, the pursuit of this goal is hindered by the limited availability of \textbf{annotated} and \textbf{publicly accessible} multimodal datasets \textbf{across a wide array of biomedical domains}. Furthermore, privacy concerns associated with sharing patient information \cite{fleming2024medalign} and the logistical complexities of expert-level annotation exacerbates 
data collection, annotation, and redistribution at scale \cite{yan2014learning}.

Scientific biomedical literature, however, provides an ever-expanding, highly curated multimodal resource encompassing the knowledge of specialized professionals, reflecting rigorously supported medical and biological evidence.
Naturally,  biomedical literature offers an unparalleled
resource to construct comprehensive datasets at scale. 

PubMed \cite{roberts2001pubmed}, an archive of biomedical and life sciences literature, indexes approximately 1.5 million new publications annually—equivalent to over two papers per minute—with a 9\% yearly growth rate.  
While prior efforts have leveraged this resource \cite{pelka2018radiology,subramanian2020medicat,lin2023pmc}, current open-source  literature-based biomedical datasets are often pre-filtered to narrow diagnostics imaging modalities within radiology and pathology, overlooking the vast breadth of complementary information available in other fields such as 
 cell and molecular biology, genetics, and pharmacogenomics  -- adjacent domains that provide insights into the very biological mechanisms that dictate health outcomes. 
Often, advancements in these areas lead to scientific discoveries that not only expand our understanding of complex biological processes but also influence medical practice (e.g. patients who metabolize codeine very rapidly are at increased risk of developing adverse effects  \cite{bertilsson2002molecular} ). Therefore, integrating knowledge from these domains is essential for effective clinical reasoning, yet it is often underappreciated in the development of biomedical vision-language training paradigms and datasets.

With the aim to democratize access to open-source scientific data across the vast landscape of biomedical research, we developed the {\bf Biomed}ical {\bf I}mage-{\bf C}aption {\bf A}rchive (\dataset), an \textbf{open-source framework} including:  an ETL pipeline to efficiently extract and serialize the entirety of PubMed Central Open Access (PMC-OA) repository into a standardized archive, as well as tools to annotate, filter, and retrieve the archive on demand. Leveraging BIOMEDICA, we present the following contributions:

\begin{itemize}
    \item \textbf{\dataset\ Dataset}: A large-scale, deep-learning-ready biomedical dataset containing over 24M image-caption pairs and 30M image-references from 6M unique open-source articles. Each data point is highly  annotated with over 27 unique metadata fields, including article-level information (e.g., license, publication title, date, PMID, keywords) and coarse-grained image metadata (e.g., primary and secondary content labels and panel type) assigned via an unsupervised algorithm and human curation by seven experts. The dataset is optimized for model development and is available in parquet \cite{vohra2016apache} format for fast  filtering and WebDataset format for high-throughput streaming (providing 3x-10x higher I/O rates when compared to random access memory). Table \ref{xc} summarizes the \dataset\ dataset, highlighting its distinctions from prior literature-based datasets.

\item \textbf{BMC-CLIP}: We continually pretrain CLIP using the \dataset\ dataset. We leverage our features and tools, such as streaming (enabling remote training) and fast filtering, to explore modern training strategies, including concept filtering \cite{fang2023data}  and data balancing \cite{alabdulmohsin2024clip}.

\item \textbf{Large-scale evaluation}: We standardized \numBenchmark\ established biomedical datasets across cell and molecular biology, radiology, pathology, ophthalmology, dermatology, and surgery to evaluate our models and compare them to prior work.  Our assessments encompass zero-shot classification, image-to-text retrieval, and text-to-image retrieval. Leveraging this benchmark, we observe that our best model surpasses previous state-of-the-art in two-thirds of the cases, achieving an average improvement of 6.56\% in general biomedical imaging classification tasks and 6.7\% in microscopy composition identification.   Furthermore, our high-throughput dataset enables our models to outperform the current state-of-the-art with 10x less compute.
\end{itemize}

\noindent Our dataset and models are hosted on Hugging Face and will be updated annually to keep pace with the growing availability of scientific literature.

\begin{figure*}[h]
    \centering
    \begin{subfigure}[b]{0.35\textwidth}
      \centering
      \includegraphics[width=\columnwidth, trim=0 0 0 0, clip]{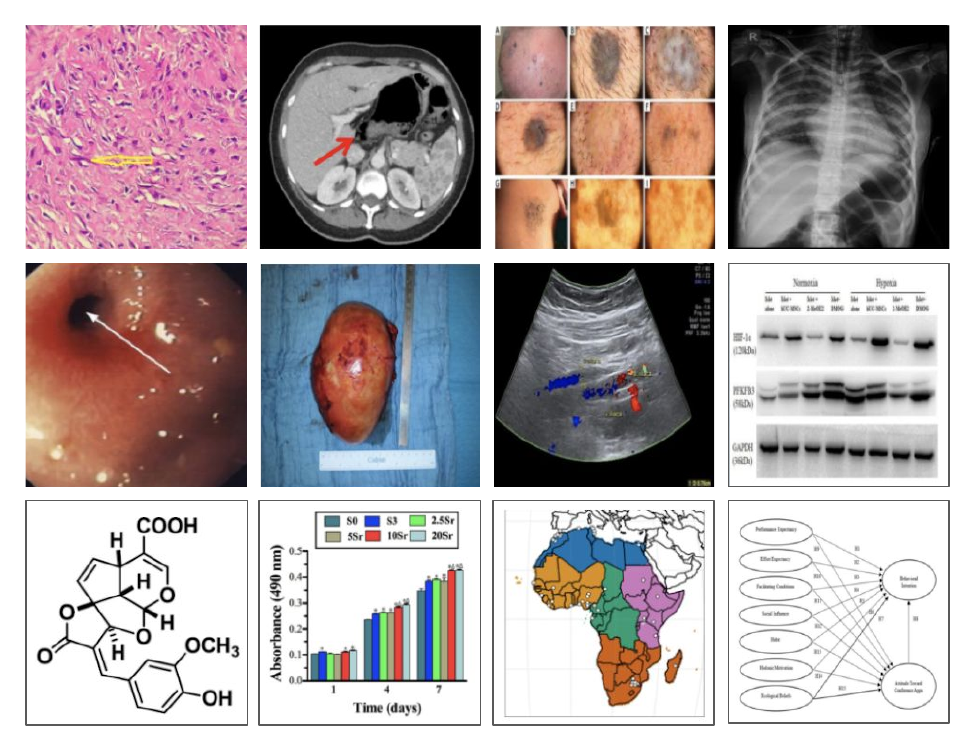}
      \label{fig:data_example}
    \end{subfigure}
    \hfill
    \begin{subfigure}[b]{0.645\textwidth}
        \centering
        \includegraphics[width=\columnwidth]{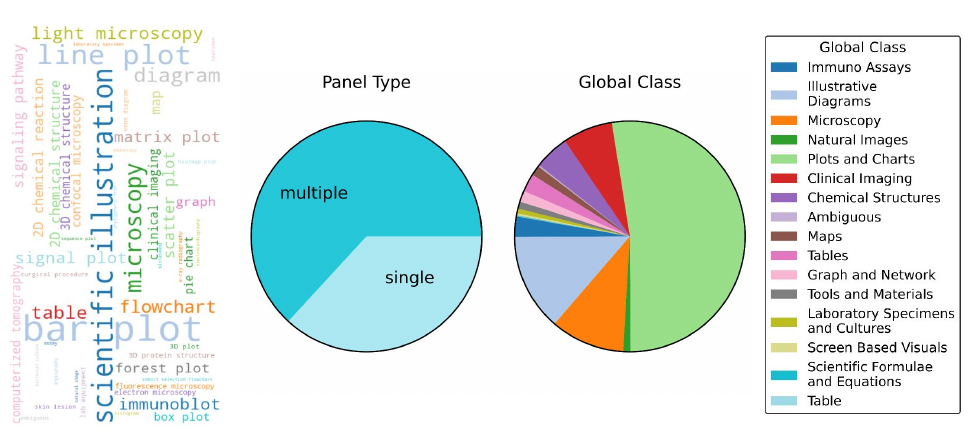}
        \label{fig:taxonomy_circle}
    \end{subfigure}
    
    \vspace{-5.5mm}
    \caption{\textbf{\textit{Left:}} Examples of images included in the \dataset\ dataset, ranging from clinical imaging to maps and bar plots. The word cloud reflects the fine-grained \textit{local concept} proportions for the most frequent concepts in the \dataset\ dataset . \textbf{\textit{Right:}} Visualization of the concept breakdown in the BIOMEDICA taxonomy. The left pie chart reflects the \textit{panel type} (light blue indicates single panel images, dark blue indicates multi panel images) and the pie chart on the right  shows the \textit{global concept} of individual taxonomies. }
    \label{fig:full_biomedica}
\end{figure*}

\vspace{-0.5em}
\section{Related Work}

\textbf{Literature-based Biomedical Image-caption Pretraining Datasets}
Biomedical image-caption pretraining mixtures leverage scientific literature to create pre-filtered  datasets. Pelka et al. \cite{pelka2018radiology} leveraged PMC-OA subset \cite{roberts2001pubmed} to develop the Radiology Objects in Context (ROCO) dataset. It filters PMC-OA images using a supervised CNN  to classify and select radiology-specifc images. This results in a collection of 81,000 image-caption pairs spanning seven radiology modalities, with additional annotations including  keywords, UMLS identifiers, and semantic types. MEDICAT \cite{subramanian2020medicat} extends ROCO by further filtering more medical images through a two-step process. First, a keyword filter searches for modality words (e.g., MRI, ultrasound) to identify medical images across captions and reference text. Next, a supervised ResNet-50 discards non-medical images. This results in 217,000 caption-figure pairs from 131,000 open-access biomedical papers. PMC-OA dataset \cite{lin2023pmc} follows the same two-step filtering strategy as MEDICAT but replaces the ResNet-50 with a ResNet-101 model trained on the DocFigure dataset \cite{jobin2019docfigure}. This approach yields 381,096 unique medical figures from 2.4 million articles, which are further refined into 1.65 million image-caption pairs.  
PMC-15M \cite{zhang2023biomedclip} scales literature-based biomedical datasets to 3 million articles, extracting 15 million image-caption pairs using PubMed Parser \cite{achakulvisut2020pubmed}.

In contrast to prior work, (1) we adopt a domain-agnostic approach. Rather than filtering to specific domains, we provide 9x more metadata and expert-derived annotations at various granularities. Subsequently, we offer a parallelized pipeline  to  use these metadata and filter subsets on demand, accommodating to different interests in the biomedical community. (2) Prior  efforts  rely on supervised models to classify images. An approach confounded by the diversity of the training datasets.  Instead, we employ a multi-step, data-driven strategy guided by clinicians and scientists to develop a comprehensive hierarchical taxonomy derived from biomedical ontologies and PMC-OA content, annotate clusters of images following this taxonomy, and  propagate these annotations  to each individual image. This data-centric approach ensures scalability, allowing newly added images to inherit labels from the nearest annotated data.

\noindent \textbf{Literature-based Biomedical Contrastive Pretraining} Large-scale, literature-based datasets enable self-supervised learning (SSL). PMC-CLIP \cite{eslami2023pubmedclip} leverages one million image-caption pairs from PMC-OA to train an image-captioning model with a dual contrastive learning and masked language modeling objective. BiomedCLIP  \cite{zhang2023biomedclip} expands on PMC-CLIP by pertaining on 15 million image-caption pairs from PMC-15M and upgrading the vision encoder to vision transformer (ViT-B-16). 

Unlike the original CLIP model \cite{radford2021learning}, prior work in the biomedical literature aligns vision-text representations from scratch with at least 8x smaller batch sizes and 26x smaller datasets -- factors linked to suboptimal model performance \cite{cherti2023reproducible, zhai2023sigmoid, li2024scaling}. Recognizing that even at 24M image-caption pairs, biomedical datasets are orders of magnitude smaller than general vision-language datasets (e.g., LAION-5B \cite{schuhmann2022laion}) and the scarcity of large compute clusters in academia, we instead investigate continual pretraining. This setup, combined with our framework, enables the exploration of modern training techniques such as data filtering, data balancing, and robust fine-tuning \cite{wortsman2022robust}.

\noindent \textbf{Evaluation of Biomedical Image-caption Embedding Models}
While biomedical CLIP-style models may be trained on diverse datasets, their evaluation is typically limited to tasks within radiology and pathology, overlooking the broader biomedical domain. In general, CLIP-style models are evaluated on retrieval, classification, and open vision-question answering (VQA) tasks. BiomedCLIP evaluates retrieval performance on a 726k held-out set from PMC-15M, with classification capabilities tested on LC25000 \cite{borkowski2019lung}, PathCamelyon \cite{Veeling2018-qh}, TCGA-TIL, and the RSNA 2018 Challenge \cite{shih2019augmenting}. PMC-CLIP assesses retrieval performance on ROCO \cite{pelka2018radiology} and classification performance on MedMNIST \cite{yang2023medmnist} (including PneumoniaMNIST, BreastMNIST, and DermaMNIST). Both BiomedCLIP and PMC-CLIP also evaluate open-VQA performance on the VQA-RAD and Slake datasets. However, to support open-ended evaluation, additional models are included, requiring full training. BiomedCLIP introduces a transformer-based co-attention multimodal module that generates cross-modal representations from image and text encodings, which are then passed to a classifier to predict the final answer. In contrast, PMC-CLIP uses Model-Agnostic Meta-Learning (MAML) networks for its approach.

These existing evaluation protocols have several limitations.  First, evaluations are narrow in scope as performance is quantified on a limited range of biomedical domains, mainly radiology and pathology. 
Second, the open-VQA evaluation of multimodal embedding models  requires training at least one additional decoder using non-standardized frameworks, leading to incomparable results. Moreover, the introduction of extra learnable parameters on top of CLIP-style models compromises the reliability of the assessment. For example, while BiomedCLIP and PMC-CLIP show at least a 40\% performance gap across all classification and retrieval tasks, this gap narrows to just 7\% in Open-VQA, suggesting that  (all things equal) the added decoders are capturing much of the information needed to solve the proposed tasks, as supposed to the evaluated encoder models. 

As an initial step in addressing these issues, we conduct a systematic assessment of zero-shot performance across \numBenchmark\ datasets, incorporating existing evaluations as thoroughly as possible, given data availability and reproducibility constraints. We include additional datasets to encompass previously overlooked biomedical domains. Lastly, all models are evaluated using the same framework, ensuring a fair and consistent comparison.

\vspace{-0.5em}
\section{\dataset\ Data Curation Process}
Our dataset curation workflow is illustrated in Figure \ref{fig:data_workflow}, and consists of three stages: (1) dataset extraction, (2) concept labeling, and (3) dataset serialization.

\subsection{Data Extraction}
Collected article data and metadata is stored in JSONL files. Table \ref{tab:metadata-list} lists all the extracted data along with its provenance.

\noindent {\bf Media file download} Compressed files containing articles (stored as nXML files) and media files are downloaded from the National Center for Biotechnology Information (NCBI) via the File Transfer Protocol service \cite{postel1985file,pmc_open_access_subset} . 
Article data and metadata is aggregated from the retrieved file list, nXML files and the Entrez API. 
\begin{itemize}
    \item {\bf File List}: Contains metadata fields to link articles to their corresponding media files. We iterate over  this index, collecting PMID (a unique accession ID), file paths, publication date, references, journal and license. 
      
    \item {\bf nXML file}: Articles are provided in a structured nXML file. By parsing the nXML, we extract article data, including:  title, abstract, keywords, category,  and  full text.
   
    \item {\bf Entrez API}: The Entrez API provides access to NCBI’s full collection of biomedical literature metadata, allowing the retrieval of  additional information. Through this API, we collect MeSH terms and PMIDs of  citing articles.
\end{itemize}

\noindent {\bf Figure Caption}
Each figure is matched with its corresponding caption by finding the graphic element in the nXML tree with an \texttt{@xlink:href} attribute matching the image id.

\noindent {\bf Figure Mentions}
In addition to the captions, we extract figure mentions—paragraphs in the article text that reference a given figure.  
To this end, we search for all XML tags with figure attribute \texttt{<xref ref-type="fig" ...>} and store all matched paragraphs. 

\noindent {\bf License}: Each image-caption pair entry includes the associated license information and the article citation. We adhere to all license terms provided by NCBI. 
To maintain strict compliance, we followed the three primary license groups—commercial, non-commercial, and other.

\subsection{Concept Labeling}
\label{cls_}

\noindent \textbf{Feature Clustering} 
All images are embedded using DINO-v2  \cite{caron2021emerging} (ViT-L-14 distilled) and  further reduced to 25 principal components (refer to Figure \ref{fig:DinoPCA}) using PCA. Subsequently, the compacted features are over-clustered using K-means with $K=2000$.

\noindent \textbf{Concept taxonomy}
A team of two licensed clinicians (pathology and surgery) and a bioinformatician were tasked with developing a hierarchical taxonomy for concepts within PMC-OA. 
First, an initial taxonomy is derived from  biomedical ontologies including Uberon \cite{mungall2012uberon}, NCBITaxon \cite{dougan2014ncbi}, and the Cell Line Ontology \cite{sarntivijai2014clo}. Subsequently,  30 samples are randomly sampled from each cluster.  Annotators are then tasked to review the image content of these clusters to  expand the taxonomy, refining it over three iterative rounds. The  taxonomy is presented in figure \ref{taxonomy-example}.

\noindent \textbf{Cluster Annotation} A group of seven individuals with expertise in genetics, pathology, surgery, developmental biology, and biomedical informatics were tasked to annotate each cluster using the generated taxonomy. 
Using the majority of observed concepts within a cluster (ordered by frequency), annotators were instructed to assign one or multiple global and local concept labels, determine whether the images were multi-panel, and, if so, specify the types of sub-panels included (see Section \ref{a:concept-labeler}).

\noindent \textbf{Label Resolution and Propagation}
A majority voting system is employed to determine all concept labels within a cluster (see Section \ref{a:concept-labeler}). In the case a concept is only identified by one reviewer, labels are re-evaluated to guarantee that all concepts are included. Once labels are assigned, a dictionary linking cluster IDs to label metadata is used to propagate the annotations to individual images.

\subsection{Data Serialization}
 Table \ref{tab:serialziation} shows total compute time for parallelized data  serialization. We convert the annotated paired media-JSON files into WebDataset format (following OpenClip's naming convention for images and captions). This structure ensures that the dataset can be efficiently processed and distributed on demand via streaming, improving scalability.

\noindent \textbf{Dataset Access} To facilitate  efficient user access, we make the \dataset\ dataset available on Hugging Face. 
This enables users to stream  our large dataset directly from the Hub without downloading the entire artifact and provides seamless access even with limited local random access memory through memory mapping.

\section{\dataset\ Dataset Description}

\noindent {\bf Downloaded Articles} A total of \numArticles\ articles are downloaded from the NCBI server through FTP. Within this collection, 5,050,473 articles have at least one image, while 992,021 articles are text-only. 
All articles have a corresponding nXML that contains the full text. 
Example images are shown in Figure \ref{fig:full_biomedica}.

\noindent {\bf Image-caption pairs collected}
From the full-text articles and associated image files, we collected a total of \numPairs\ unique image-caption pairs and extracted 30,711,542 figure references. 
On average, each article contains 4.9 images and each image is accompanied by 1.6 figure references.

Table \ref{tab:data-statistics} presents detailed text and image statistics. Caption token lengths range from a single token to 12,389 tokens, with a median length of 64 tokens, indicating that most captions are concise but can vary substantially. 
In contrast, figure reference paragraphs—which describe or contextualize images within the article—tend to be longer, with token counts reaching up to 699,117 tokens and a median of 338 tokens, showing the level of detail often required for clinical or scientific context.
Overall, the text content includes a total of $2.84 \times 10^9$ tokens for captions, and $6.65 \times 10^{10}$ tokens for the full text, illustrating the extensive scale of language data within the dataset.

For images, the median image width and height are 709 and 476 pixels, respectively. However, the wide range in both dimensions, from pixel up to tens of thousands (with a maximum width of 52,490 pixels and height of 65,081 pixels). This variability is due to the presence of both thumbnails or low-resolution images and high-resolution images, such as full-page figures or detailed illustrations.

\noindent {\bf Image taxonomy} The image taxonomy comprises 16 global concepts and 119 local concepts (see Figure \ref{fig:full_biomedica}). Table \ref{tab:global_taxonomy} shows image counts by global concept. 
Biomedical (microscopy and clinical imaging) images represent 17\% of the total dataset. Notably, biomedical images are more common in noncommercial publications, where they make up 13\% of the images, compared to 5\% and 6\% in commercial and other license categories, respectively. 
The ``Plots and Charts" category holds the largest count across all source types, with a total of over 13 million images, which is 57\% of the images, demonstrating that graphical data representations are common across scientific literature.

\begin{table*}[t]
\centering
\small  
\begin{adjustbox}{width=\textwidth}
\begin{tabular}{p{3.5cm}C{1.2cm}C{1.2cm}C{1.4cm}C{1.4cm}C{1.5cm}C{1.5cm}C{1.5cm}C{1.9cm}C{1.6cm}C{1.5cm}C{1cm}}
\toprule
\rowcolor{gray!10} \textbf{Model} & \multicolumn{2}{c}{\textbf{Biology}} & \multicolumn{3}{c}{\textbf{Pathology}} & \multicolumn{2}{c}{\textbf{Radiology}}  & \textbf{Ophthamology} & \textbf{Dermatology} & \textbf{Surgery} & \textbf{Average}  \\

\cmidrule(lr){2-3} \cmidrule(lr){4-6} \cmidrule(lr){7-8}  

& \makecell{Cell \\ Profiling} & \makecell{Structure \\ Profiling} & \makecell{Cytology} & \makecell{Neo- \\Histopath} & \makecell{Non Neo- \\Histopath} & \makecell{Chest \\ X-ray}  & \makecell{Breast \\ ultrasound} & \makecell{Fundus \\ Camera} & \makecell{Skin \\ diagnosis} & \makecell{ M.I. \\ surgery} & \makecell{ Total}  \\

\rowcolor{gray!10} \textit{Num tasks $  \lfloor $ Avg. Options $  \rfloor $ } & 6 (5)  & 3 (7) & 3 (6) &  4 (5) & 4 (3)  & 2 (3) & 1 (2) & 1 (4) & 1 (7)& 10* (2) & 35 (4) \\
\midrule
Random & 20.00	& 14.28 & 16.66	& 20.00  & 33.33  & 33.33  & 50.00  & 25.00 & 28.00 & 50.00 & 29.06 \\
\midrule
OpenCLIP \cite{ilharco_gabriel_2021_5143773} &	24.63 &	44.84 &	\underline{20.24} &	43.88 &	36.81 & 60.88 & \underline{68.27} & 23.75 & 20.37 & \underline{67.51} & 41.18  \\
CoCa \cite{yu2022coca}	& 23.58	& 32.92	& \textbf{36.27}& 36.50 &	38.33	& 24.69	& 30.45 &	43.50&	10.15  & 36.78 & 31.12\\

\midrule
PMC-CLIP \cite{lin2023pmc} & 6.01& 11.23	& 12.48 & 29.86 & 7.82 &  28.47 &	58.97 & 20.12 & 12.59 & 42.96  & 23.05 \\
BioMedCLIP  \cite{zhang2023biomedclip} & \underline{26.54} & 49.14 & 17.03 & 48.83 & \textbf{42.39}& 56.04	 & 56.09 & 26.12 & 36.01 & 53.40 & 41.16 \\

\midrule

\rowcolor{med!10} \modelMed  & 24.36 & \textbf{	53.51}& 19.48 & 	\textbf{55.81}& 	38.60& 	56.68	& 64.10	& \textbf{43.62}& \textbf{65.81}& 	55.27	& \textbf{47.72} \\

\rowcolor{med!10} \modelBal &  23.54	&50.35	&16.39	&\underline{52.33}	&33.62	&50.34&	66.35	&\underline{43.50}	&\underline{56.01}&	56.51	&44.89\\

\rowcolor{gen!10} \modelGen &  23.56	&48.66	&16.61&	51.97&	34.55	&53.06&	56.41&	43.00	&23.07&	62.64&	41.35\\

\rowcolor{med!10} \modelMedM  & 26.27& 	\underline{51.68}&	16.67	&49.08&	\underline{39.07}&	\textbf{71.50}&	67.31&	37.88	&28.45&	62.18	& \underline{45.01}  \\

\rowcolor{bal!10} \modelBalM  & 24.66 &	50.44	&17.19&	49.09&	38.50	&64.59&	\textbf{70.19}	&36.75	&24.06	&62.39&43.79 \\
\rowcolor{gen!10} \modelGenM	&\textbf{ 27.70}&	47.67	&16.82	&44.19	&38.62	&\underline{66.32}&	\underline{68.27}&	27.62&	15.84&	\textbf{67.68}	&42.07 \\
\bottomrule
\end{tabular}
\end{adjustbox}
\caption{ Average zero-shot classification performance across 35 classification tasks (from 21 unique datasets) stratified by domain and task. We show results for three variants of our model, including continual pretraining on all of \dataset, a subset after concept-filtering (CF), and a subset after concept-balancing (CB). Models with indication WiSE-FT are merged counterparts as described in \cite{wortsman2022robust}. \textbf{Bold} indicates best performance, \underline{underline} indicates second best performance.
}
\label{tab:zero_shot}
\end{table*}


\section{Evaluation Benchmark}
To evaluate the effectiveness of continual pretraining  (further training a pretrained model on new data) on the \dataset\ dataset, we repurposed 39 established biomedical classification tasks and leveraged a new retrieval dataset based on Flicker, for a total of 40 datasets.

\noindent \textbf{Image classification task selection}
To construct a comprehensive classification benchmark, we take the union of evaluations from prior work (BioMedCLIP and PMC-CLIP) and supplement it with underrepresented domains. For each individual task, classes are converted into captions (see Supplements), providing two variations per class. This evaluation set spans multiple fields: pathology (11 tasks), radiology (3 tasks), ophthalmology (1 task), dermatology (1 task), surgery (10 tasks),  biology (9 tasks) and general microscopy (4 tasks). The biology and pathology tasks are sourced from Micro-Bench \cite{lozano2024mu}; ophthalmology and dermatology tasks from MedMnist \cite{medmnst}; radiology tasks from RSNA \cite{shih2019augmenting} and CheXpert \cite{chexpert}; and minimal invasive (M.I.) surgery tasks are collected from the Dresden surgical  dataset \cite{carstens2023dresden}. Table \ref{table:benchmark_sources} provides a brief data description, citation, and domain.

The image classification benchmark is subdivided into two splits.  The first category, general bioimaging classification, encompasses tasks such as imaged-based  medical diagnosis, object identification,  and cellular profiling (e.g.  cell state classificat).  This subset covers 35 tasks.
The second split, microscopy composition identification, is derived from the Micro-Bench Perception coarse-grained benchmark. It spans both pathology and biology across  light, fluorescence, and electron microscopy. These tasks involve identifying basic micrograph properties, including:
The domain (DM) or field of the image  (e.g. pathology, the microscopy modality (MD) and submodality (SM) used for image acquisition (e.g. light microscopy), and the staining technique (ST) applied to the specimen (e.g. H\&E).

\begin{figure}
    \centering
    \includegraphics[width=1\linewidth , trim=10 35 10 10]{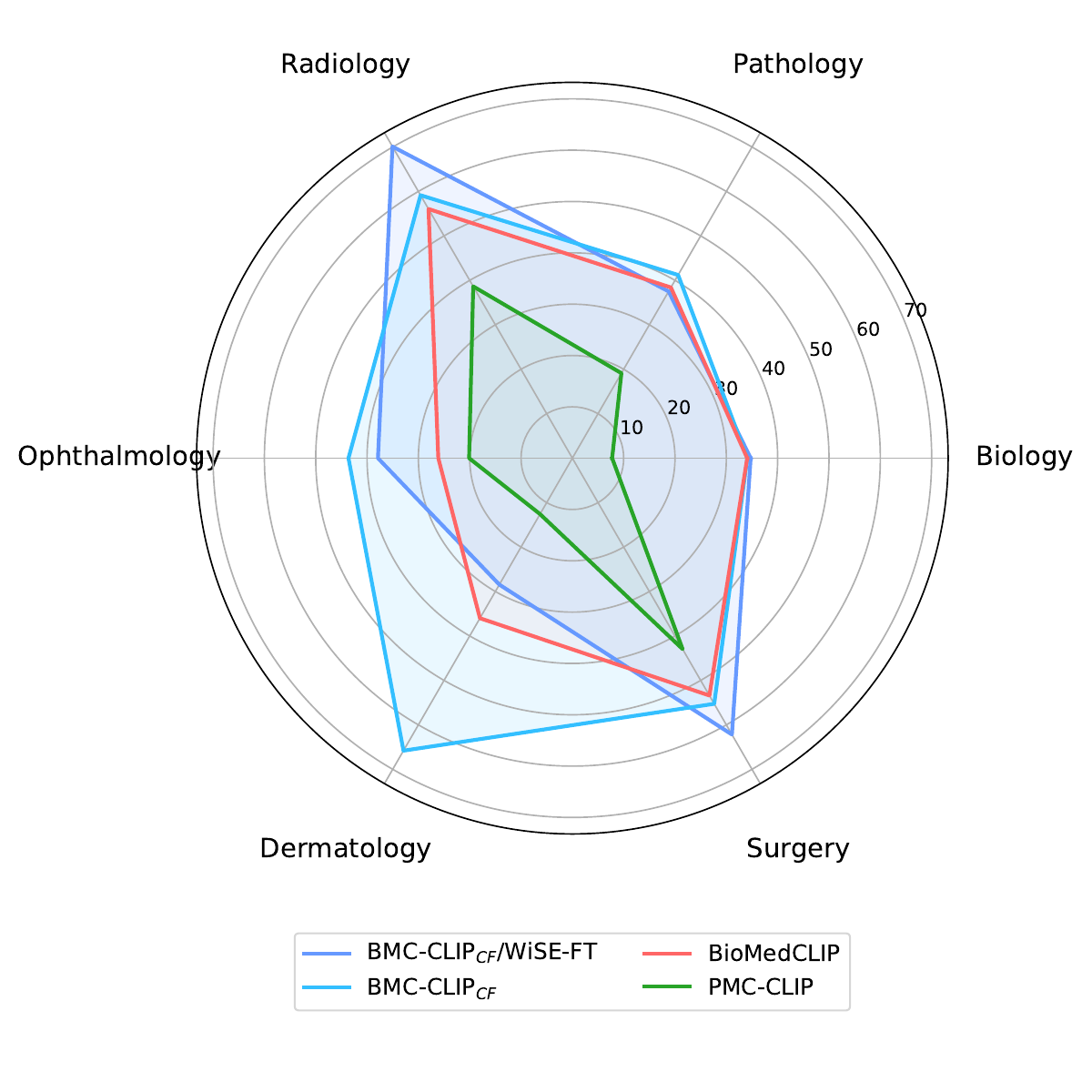}
    \vspace{-2em}
    \caption{Average model performance across different biomedical domains of best BMC-CLIP models compared to prior work.}
    \label{fig:radial_plot}
\end{figure}

\noindent \textbf{Multimodal retrieval task selection}
We assess multimodal retrieval performance using a new collection of 7K high-quality, open-source biomedical image-caption pairs from Flickr. This benchmark spans concepts across pathology, radiology, biology, dermatology, and surgery (see Supplements Section \ref{a:flickr_data} for a dataset description and samples).

\noindent \textbf{Metrics.} Classification tasks are evaluated using average accuracy across two caption variations. Retrieval tasks are measured using recall at 1, 10, and 100.

\section{Experiments}

We examine the effects of (1) continual pretraining (using the information noise-contrastive estimation loss \cite{oord2018representation}) on the full set of 24M image-caption pairs and compare these insights against (2) topic balancing, (3) dataset filtering, (4) robust fine-tuning. 
This topic exploration is non-exhaustive and it is designed to utilize the supplementary annotations, metadata, and features made available in our dataset.

\begin{table}[h!]
\small
\centering
\resizebox{\columnwidth}{!}{%
\begin{tabular}{lp{0.8cm}p{0.8cm}p{0.8cm}p{0.8cm}p{0.8cm}p{0.8cm}}
\toprule
\rowcolor{gray!10} \textbf{Model} & \textbf{DM} & \textbf{MD} & \textbf{ST} & \textbf{SMD}  & \textbf{Avg.}\\
OpenCLIP \cite{ilharco_gabriel_2021_5143773} & 49.67 & 85.66 & 46.77 & \underline{29.23}& 53.03  \\
CoCa \cite{yu2022coca} &	47.29	& 71.89& 	39.27 & 	\textbf{48.00} & 47.29\\

\midrule

PMC-CLIP \cite{lin2023pmc}& 5.56 & 14.68 & 10.92 & 12.49 & 10.91 \\
BioMedCLIP \cite{zhang2023biomedclip} & 41.54 & 74.72 & 49.74 & 32.89  & 49.72\\
\midrule
\rowcolor{med!10} \modelMed	& 49.07	& 78.71& 	59.17&	14.54	&50.37 \\

\rowcolor{bal!10} \modelBal & 40.07	& 83.4 &  57.28 & 	25.48 &  	51.55\\

\rowcolor{gen!10} \modelGen & 	44.56& 	85.70& 	58.40	& 26.72 & 53.84 \\

\rowcolor{med!10} \modelMedM	&\textbf{58.78}  &	\underline{86.88} &	\underline{60.84}  & 27.65  & \textbf{58.53}\\

\rowcolor{bal!10} \modelBalM &	52.12 &	\textbf{87.21} &	\textbf{61.84} & 25.04 & 	\underline{56.55}\\

\rowcolor{gen!10} \modelGenM &	 \underline{56.26}	&83.97 &54.81 &	27.89 &	55.73\\

\bottomrule
\end{tabular}
}
\caption{Microscopy composition identification performance in 4 course-grained classification tasks across pathology and biology. \textbf{Bold} indicates best performance, \underline{underline} indicates second best.}
\vspace{-1em}
\label{course_grained}
\end{table}


For all experiments, we continually pretrain OpenCLIP (ViT-L-14) using a batch size of 1024 per GPU (distributed across four H100 80GB GPUs) with a batch accumulation frequency of two steps, resulting in an effective batch size of 8192 samples per step. We use a learning rate of 1e-6 with 1K warmup, and 32-bit floating-point precision. All of our experiments are trained via streaming, eliminating the need for local storage of the \rawDataSize TB dataset. Additional training details  are shared in supplements section \ref{a:model-training}. With these configurations, we run the following experiments:

\begin{enumerate}

\item \textbf{Continual pretaining on full dataset (24M)}
In this experiment, we continually pretrain OpenCLIP   on the complete \dataset\ dataset (24M data points). We train this model for 9 epochs. This experiment serves as a baseline  to compare additional data mixture strategies. 

\item  \textbf{Concept-Balancing (8M)}
For this experiment, we continually pretrain  OpenCLIP   on 8M image-caption pairs obtained by balancing the frequncy of local topics from the \dataset\ dataset. To this end, over represented topics (e.g. plots) are dropped.  This model is trained  for 27 epochs. This experiment targets potential biases introduced by data imbalance by restricting category over-representation. 

\item \textbf{Concept-Filtering (6M)}

In this experiment, we  filter the  \dataset\ dataset, only  retaining concepts within clinical \&  scientific imaging, immunoassays, illustrative diagrams, chemical structures, maps, tools \& materials, and hand-drawn/screen-based visuals (excluding tables, figures, and scientific equations).  We then use this filtered dataset to continually pretrain OpenCLIP for 36 epochs.  

\item  \textbf{Robust Fine-tuning}
This experiment explores model merging  \cite{wortsman2022robust} by using a convex combination of the weights from a base model and its adapted counterpart. 

\end{enumerate}

\section{Results}

\begin{table*}[ht]
\small
\centering
\begin{adjustbox}{max width=\textwidth}
\begin{tabular}{lcccccc}
\toprule
\rowcolor{gray!10} \textbf{Model} & \multicolumn{3}{c}{\textbf{Image $\rightarrow$ Text}} & \multicolumn{3}{c}{\textbf{Text $\rightarrow$ Image}} \\ 
\cmidrule(lr){2-4} \cmidrule(lr){5-7}
               & Recall@1 & Recall@10  & Recall@100 & Recall@1 & Recall@10  & Recall@100 \\ \hline

OpenCLIP \cite{ilharco_gabriel_2021_5143773}  &  2.78 & 9.78 &26.75 & 2.91 & 9.51 &24.59\\

CoCA  \cite{yu2022coca}        & 1.68 &	5.47 &	15.96&	1.50 &5.51&	15.24 
 \\


\midrule

PMC-CLIP  \cite{lin2023pmc}   & 0.03 & 0.13 & 1.39 & 0.00 & 0.13 &1.50 \\   
BiomedCLIP  \cite{zhang2023biomedclip} & 3.70 & 12.78 & \underline{36.27} & \underline{3.94} & \underline{13.63} & \underline{35.63} \\

\midrule

\rowcolor{med!10} \modelMed  &\textbf{ 4.13} &\textbf{ 15.13 }& \textbf{38.30 } & \textbf{4.15 }& \textbf{13.75} & \textbf{36.10}\\

\rowcolor{bal!10} \modelBal & 3.90 & 13.24 & 33.43 & 3.66 & 12.22 & 31.80 \\

\rowcolor{gen!10} \modelGen &  3.67 & 12.72 &  31.93 &  3.41 &  11.20 &  30.03 \\

\rowcolor{med!10} \modelMedM & \underline{3.90} &  13.24 &  33.43 & 3.66 & 12.22 &  31.80\\

\rowcolor{bal!10} \modelBalM & 3.97 & \underline{13.57} &33.32 & 3.58 & 12.43 & 31.82\\

\rowcolor{gen!10} \modelGenM & 3.83  &12.68 & 30.94  & 3.26 & 11.57 & 29.19
\\

\bottomrule
\end{tabular}
\end{adjustbox}
\caption{Top-K retrieval performance on BioMed-Flickr. \ \textbf{Bold} indicates best performance, \underline{underline} indicates second best performance.}
\label{table:retrieval_performance}
\end{table*}

\textbf{Concept Filtering leads to better performance across zero-shot classification and retrieval tasks}
Intuitively, when compared to other continual pretraining strategies within \modelGen\ models, filtering the dataset (e.g., dropping over-represented topics like plots and tables) yields the best average performance across general biomedical imaging classification tasks with respect to concept balancing (better 80\% of the time) and full-dataset pretraining (better 90\% of the time). Additionally, concept filtering leads to better performance compared to concept balancing or full-dataset pretraining in image-to-text and text-to-image retrieval.
Indeed, within this training strategy, 48\% of the data mixture corresponds to clinical imaging and microscopy.  These results align with those observed by DataComp \cite{gadre2024datacomp}. While scaling boosts performance, filtering pretraining data to match evaluation domains provides additional gains over unfiltered data.

\noindent \textbf{Models Trained on the \dataset\ Dataset Lead to State-of-the-Art Zero-Shot Performance}
Compared to prior work, models trained on the \dataset\ dataset yield better average performance in classification and retrieval tasks. \modelMed\ outperforms PMC-CLIP in all tasks, achieving a +24.67\% improvement in general biomedical imaging classification tasks, with a minimum gap of +5.13\% in ultrasound (radiology) and a maximum gap of +53.22\% in dermatology. Similarly, a +39.46\% improvement is observed in microscopy composition tasks. Additionally, a recall@100 gap of +36.91\% and +34.6\% is observed in image-to-text and text-to-image retrieval, respectively. Similarly, \modelMed\ outperforms BioMedCLIP in 8/10 general biomedical imaging classification subsets. yielding an average improvement of 6.56\%. For individual tasks, \modelMed\ achieves  the highest differential performance w.r.t BioMedCLIP  in  dermatology (+29.8\%),  ophthalmology (+17.5\%), breast ultrasound (+8.01\%) and  non-neo histopathology (+6.98 \%) and marginal better performance in microscopy composition identification and all retrial evaluations. It is noteworthy to highlight that \modelMed\ achieves these results while using 10× less compute and 2.5× less data than BioMedCLIP.

\noindent \textbf{Robust Fine-tuning complements model performance in a subset of tasks} Another advantage of continual pretraining is the capability to improve model performance without further training via  merging \cite{wortsman2022robust,lozano2024mu}. 
For example, in microscopy composition identification tasks, WiSE-FT improves the  performance of \modelMed\  by 8.16\%, further increasing the accuracy gap with respect to BioMedCLIP. Similarly, WiSE-FT enhances  \modelMed 's performance in 5/10 general biomedical imaging classification tasks (Figure \ref{fig:radial_plot}). Notably, \modelMedM\ increases the performance gap w.r.t BioMedCLIP in x-ray radiology (+15.46\%), ultrasound (+11.22\%), and surgery (+8.78\%) image classification, complementing weakness in the original \modelMed\ model. However, these gains come at the cost of lower performance in other tasks, such as decrease of  -7.56\% in dermatology image classification  and marginally worse multimodal retrieval recall.

\vspace{1em}

\section{Limitations}
While our models offer state-of-the-art performance, all evaluations indicate that there is still significant room for improvement. We highlight two promising directions:
{\bf Short context length}: CLIP (ViT-L-14) has a maximum context length of 77 tokens, which restricts our ability to fully utilize captions exceeding this length. While the median caption length in our dataset is 64 tokens, there remains a portion of longer captions that cannot be leveraged  during training. This limitation may lead to information loss potentially impacting model performance on tasks requiring more comprehensive textual understanding. 

\noindent {\bf Varied image size}: The \dataset\ dataset contains images of diverse sizes and resolutions, reflective of the original PMC-OA formats. For model training, all images are uniformly resized to meet the input requirements of the image encoder, without any preprocessing to address the variation in image quality or aspect ratios. Thus, high-resolution images with fine-grained details may lose important visual information when downscaled, while low-resolution images may become blurry when upscaled, distorting features and potentially introducing misleading information.

\section{Conclusion} 
In this work, we present \dataset, a framework for converting  scientific literature (PMC-OA) into the largest deep-learning-ready dataset, comprising 24 million image-caption pairs with 27 metadata fields. We demonstrate the utility of the \dataset\ dataset by continually pretraining CLIP-style models, fully leveraging its expert-guided annotations, metadata, and streaming capabilities. Our results showcase the effectiveness of this resource, even in low-memory and GPU-constrained scenarios. Our models achieve state-of-the-art zero-shot classification performance using prior open-source tools and base models, while utilizing 10x less compute and 2.5x less data—underscoring the importance of large-scale annotated open datasets. The \dataset\ framework, dataset, models, and large-scale evaluation serve as a foundation for advancing vision-language research and applications in scientific and biomedical domains. We release all our contributions under a permissive license to facilitate broader use and further development.

{
    \small
    \bibliographystyle{ieeenat_fullname}
    \bibliography{main}
}


\clearpage
\setcounter{page}{1}
\maketitlesupplementary

\newpage
\suppfigure 
\supptable  

\section* {Supplementary Material Table of Content}
\begin{enumerate}

    \setcounter{enumi}{9}

    \item Acknowledgments
    
    \item Dataset Description

    \item Dataset Statistics
     
    \item Dataset Curation Process
    \subitem 13.1. PMC OA Dataset Description
    \subitem 13.2. Data Extraction
    \subitem 13.3. Dataset Serialization
    \subitem 13.4. Tokenized Caption Distribution
    
    \item Concept Labeling: Additional Details
     \subitem 14.1. Dimensionality Reduction
     \subitem 14.2. Over-Clustering
     \subitem 14.3. Online Cluster Annotation Form     
     \subitem 14.4. Taxonomy Curators Statistics
     \subitem 14.5. Cluster Annotator Statistics
     \subitem 14.6. Dataset Taxonomy
     \subitem 14.7. Label Assignment and Propagation
     \subitem 14.8. Inter-annotator Disagreement
    
     \item Data Upload

     \item Model Training
      \subitem 16.1. Base Model Selection
      \subitem 16.2. Modeling Hyperparameters

     \item Evaluation
     \subitem 17.1. Closed VQA Benchmark
     \subsubitem 17.1.1. Closed VQA Formulation
     \subsubitem 17.1.2. Closed VQA Evaluation
     \subsubitem 17.1.3. Closed VQA Conversion Prompts
     \subitem 17.2. Retrieval Benchmark Evaluation
     \subitem 17.3. Computing Confidence Intervals 

    \item Flickr Dataset Description
    
    \item Compute Environment

\end{enumerate}


\section{Acknowledgments}

This research was supported by NIH grants (NIH\#P30AG066515 to JJN), the Chan Zuckerberg Initiative Neurodegeneration Challenge Pairs Pilot Project   to SYL (2020-221724, 5022), the Wu Tsai Knight Initiative Innovation grant (\#KIG 102) to SYL, Hoffman-Yee Research Grant to SYL, the Arc Institute Graduate Fellowship to AL, the Stanford Data Science Graduate Research Fellowship MW, and the  Quad Fellowship to JB. SYL is a Chan Zuckerberg Biohub – San Francisco Investigator. We  thank Daniel van Strien, Matthew Carrigan, and Omar Sanseviero from Hugging Face for their invaluable assistance with data upload and design planning on the Hugging Face platform.

\section{Dataset Description}

\begin{figure}[hbt]
\includegraphics[width=0.4\textwidth]{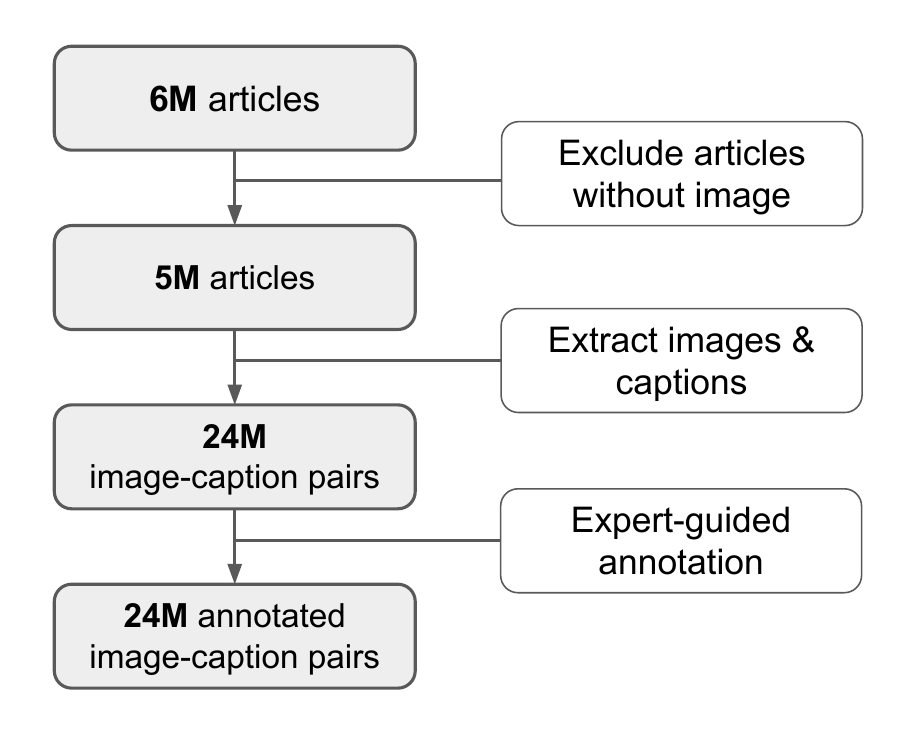}
\centering
\caption{ \dataset\ cohort diagram: selection criteria for the  construction  
of relevant image-caption pairs.  }
\label{fig:data_flow}
\end{figure}

\dataset\ dataset contains a total of \numPairs\ image-caption pairs from 5,050,473 scientific articles (full article-text for the 6M is  additionally provided). 
Each image-caption pairs  is assigned  article metadata and additional annotations. Table \ref{tbl:BiomeDicaDatasetStatistics} shows descriptive statistics for \dataset\ dataset. Table \ref{tab:metadata-list} shows all the data and  metadata  fields provided in the dataset. Table \ref{biomedica-example} shows example  image-caption pairs.   Figure \ref{fig:data_flow} shows a cohort diagram, summarizing the materialization
of \dataset\  dataset.

\begin{table}[h!]
    \centering
    \caption{\dataset\ dataset statistics: Counts of articles, image-caption pairs, metadata, human annotators and specialties.}
    \begin{tabular}{lr}
    \toprule

\rowcolor{gray!10}    \textbf{Aspect} & \textbf{Count} \\

    Articles & \numArticles \\
    Articles with Images & 5,050,473 \\
    Images & \numPairs \\
    Captions & \numPairs \\
    Figure Reference & \numFigureReference \\
    Metadata & 22 \\
    Global Concepts & 13 \\
    Local Concepts & 170 \\
    Clinical Annotators & 2\\
    Scientist Annotators & 6\\

    \bottomrule
    \end{tabular}
\label{tbl:BiomeDicaDatasetStatistics}
\end{table}


\begin{table}[]
\small
\centering
\begin{adjustbox}{width=0.55\columnwidth}
\begin{tabular}{ll}
\toprule
\textbf{Concept} & \textbf{Provenance} \\ \hline
\rowcolor{gray!10} \textbf{Image Data} & \\
Image Key & Generated  \\
Image File & File List \\ 
Caption &  nXML File \\ 
\hline
\rowcolor{gray!10} \textbf{Image Metadata} & \\
Image Cluster ID & Generated  \\ 
Image Hash & Generated \\ 
Image File Name & nXML File \\ 
Image Set &  nXML File \\ 
Image Context &   nXML File \\ 

\rowcolor{gray!10} \textbf{Image Annotations} & \\
Image Panel Type   & Generated  \\ 
Image Panel Subtype   & Generated  \\ 
Image Content Primary Label      & Generated  \\ 
Image Content Secondary Label     & Generated  \\ 

\rowcolor{gray!10} \textbf{Article Metadata} & \\

Article Keywords    &  nXML File \\
Article Category & nXML File \\ 
Article Title      &  nXML File \\ 
Article Abstract   & nXML File   \\ 
Article full text &  nXML File \\
Article Publication Date       &  Entrez  API \\ 
Article MeSH Terms & Entrez  API \\
Article Journal    & File List \\ 

Article PMID & File List \\ 
Article Citation & File List \\ 

Article License & File List  \\ 

List of PMIDs citing article & Entrez   API\\ 
Count of PMIDs citing article & Entrez API \\

\rowcolor{gray!10} \textbf{Image Embeddings} & \\
Image DINO-V2 Features  & Generated  \\ 
Image PMC-CLIP Features & Generated  \\ 
\bottomrule
\end{tabular}
\end{adjustbox}
\caption{List of image data (n=3), image metadata (n=5),  image annotations (n=4), article metadata (n=13), and image embeddings (n=2) provided in \dataset\ dataset, alongside source.}
\label{tab:metadata-list}
\end{table}

\section{Dataset Statistics}
\label{a:dataset-stats}

We provide additional statistics for our dataset. The reference count, representing the number of articles citing a given article, ranges from 0 to 3346, with a median of 37 (IQR: 35). Articles with no references account for 120,870 entries. A total of 263,836,608 references is found across the entire dataset.

For MeSH terms, the dataset includes 29,859 unique terms. MeSH term counts per article range from 0 to 44, with a median of 0 (IQR: 10) and a total of 32,896,861 terms. Notably, 2,991,141 articles are annotated with MeSH terms, emphasizing the dataset's depth in biomedical categorization.

Table \ref{tab:mesh_terms} summarizes the top 50 most frequent MeSH terms in the dataset.

\begin{table*}[h!]
\small
\centering
\begin{adjustbox}{max width=\textwidth}
\begin{tabular}{lccccc}
\toprule
\rowcolor{gray!10} \textbf{Text Statistics}          & \textbf{Min} & \textbf{Max}         & \textbf{Median} & \textbf{IQR} & \textbf{Total}        \\
Caption Token Length              & 1            & 12389                & 64              & 134          & $2.84 \times 10^9$    \\
Caption Character Length          & 1            & 25539                & 246             & 498          & $1.05 \times 10^{10}$ \\
Figure Reference Token Length     & 17           & 699117               & 338             & 299          & $1.28 \times 10^{10}$ \\
Figure Reference Character Length & 39           & 1395195              & 1323            & 1162         & $4.82 \times 10^{10}$ \\
Full Text Token Length            & 20           & 2449711              & 10306           & 8831         & $6.65 \times 10^{10}$ \\
Full Text Character Length        & 70           & 8701762              & 39072           & 32080        & $2.47 \times 10^{11}$ \\ \hline
\rowcolor{gray!10} \textbf{Image Statistics}         &              &                      &                 &              &                       \\
Image Width (pixels)              & 1            & 52490                & 709             & 150          &  -                     \\
Image Height (pixels)             & 1            & 65081                & 476             & 406          &   -                      \\
Image Area (pixels²)              & 1            & $1.95 \times 10^{9}$ & 334400          & 307272       &    -                     \\ 

\bottomrule
\end{tabular}
\end{adjustbox}
\caption{Overview of dataset statistics, detailing text token and character lengths, and image dimensions. For text statistics,  tokens are generated using the BPE  tokenizer from the \texttt{tiktoken} library}
\label{tab:data-statistics}
\end{table*}


\section{Data Curation Process }
\label{a:data-curation}

Besides releasing the code to make \dataset\ dataset fully reproducible, we provide additional descriptions and design choices for the dataset curation process. To increase efficiency and enable scaling, everything step in the data curation process is parallelized, unless specified otherwise.

\subsection{PMC OA Dataset Description}
The PubMed Central (PMC) Open Access (OA) Subset is a publicly accessible collection of full-text scholarly articles hosted by the National Center for Biotechnology Information (NCBI). 
This subset contains articles that have been made available under various open-access licenses. 
It covers a wide range of disciplines within biomedical and life sciences, providing rich content that includes research articles, reviews, case reports, and clinical trials. 
As of 2024, over six million articles are available, with tens of thousands of new articles added annually, reflecting the continuous contributions of researchers worldwide.

\begin{figure*}[htbp]
    \centering
    \includegraphics[width=0.75\linewidth]{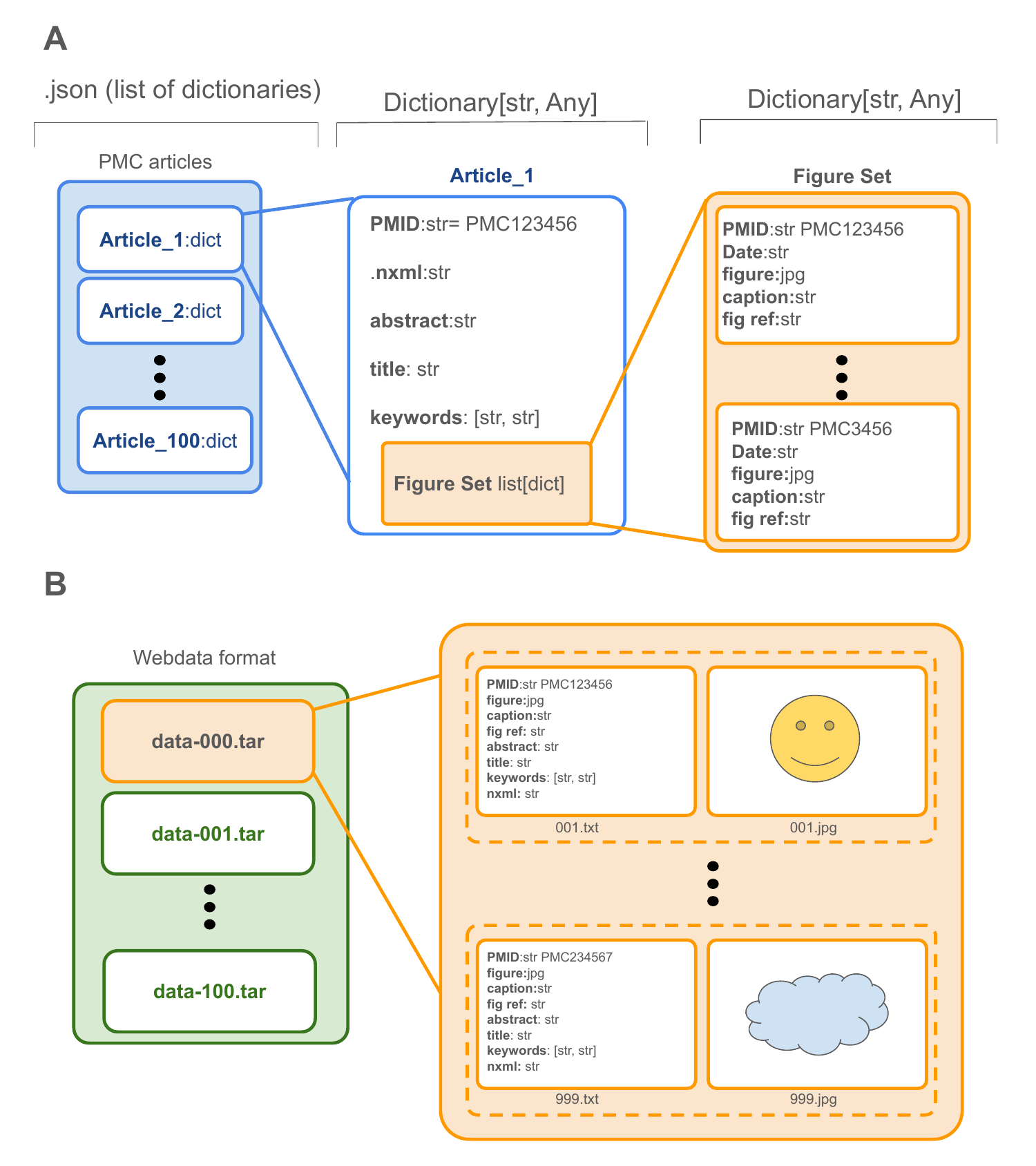}
    \caption{A) Diagram illustrating the structure of a JSON file containing a list of dictionaries representing PMC articles. Each article dictionary includes metadata fields such as PMID, nXML path, abstract, title, keywords, and a nested figure set. The figure set is a list of dictionaries, where each dictionary contains the figure's PMID, volume number, image file, caption, and context. B) Diagram illustrating the WebDataset format, where data is stored across multiple .tar archives (e.g., data-000.tar, data-001.tar). Each archive contains paired text and image files representing individual records.}
    \label{fig:json_structure}
\end{figure*}

\subsection{Data Extraction}

The remote paths to the compressed files (tar.gz) containing each article’s media files are listed in a CSV file, referred to as the file list. 
The file list provides the structure for locating and accessing the content, ensuring that all relevant files can be traced and downloaded accurately.
The file list includes the following columns: \textit{File}, \textit{Citation}, \textit{Accession\_ID}, \textit{Date}, \textit{PMID}, and \textit{License}.
The server stores files with randomized paths to optimize storage; therefore, the absolute file path provided in the file list is required to retrieve the media files for a specific article. 
We connect to the server using the Python package \texttt{ftplib}:

\begin{verbatim}
ftp = FTP("ftp.ncbi.nlm.nih.gov")
ftp.retrbinary("RETR <remote_file_path>", 
open("<local_file_path>", "wb").write)
\end{verbatim}
Bulk downloads are available only for text files, necessitating individual retrieval of associated media files.

The server enforces a rate limit of three requests per second per IP address, with varying download speeds requiring precise scheduling to prevent disruptions. 
FTP connections may disconnect intermittently, requiring a robust retry mechanism with short delays to maintain data integrity. 
To conserve storage, only necessary files (nxml and jpg) are kept, while other files (e.g., pdf, docx, xlsx, mp4) are discarded if present.

After downloading and uncompressing all files, we create JSON files to store data extracted from the raw nxml and image files. 
These JSON files contain a list of dictionaries, where each dictionary holds the data for a single unique article. 
The \texttt{figure\_set} is a list of dictionaries, where each dictionary contains the figure’s \texttt{PMID}, volume number, image file, caption, and context. 
This structure is shown in Figure~\ref{fig:json_structure}.

Entrez is a search and retrieval system from NCBI that we use to collect additional metadata, including publication details and MeSH terms, which are not available in the file list or raw nxml files. 
The Entrez API supports batch queries with up to 200 PMIDs per request:

\begin{verbatim}
from Bio import Entrez
Entrez.email = "<your_email>"
handle = Entrez.efetch(
    db = "pubmed", 
    id = "<comma_separated_PMIDs>", 
    retmode = "xml")
\end{verbatim}

Each JSON file is limited to a maximum of 200 articles to comply with the Entrez API batch limit and to keep file sizes manageable for processing.

\subsection{Dataset Serialization}
As shown in Figure \ref{fig:json_structure} A, the retrieved data is a collection of articles with full metadata and a figure set (containing multiple images and captions). This structure can natively be serialized by article; however, it requires an extra iteration within the figure set. Instead, we decide to serialize the dataset by figure, such that a row  (figure within a figure set) becomes a unique image-caption pair rather than an article. This implies that each image-caption includes all the corresponding metadata. Note that this comes with the disadvantage of repeated entries for images belonging to the same figure set (e.g. within the same publication). In other words, if two different images come from the same article figure set, then all the metadata and nxml will be repeated twice. Figure \ref{fig:json_structure} shows the data structure before (A) and after (B) serialization.

\vspace{1em}

\noindent PMC-OA Subsets were serialized in parallel. Table \ref{tab:serialziation} shows the total serialization run time. Notably, we can serialize the entirety of PMC-OA within a single day.

\begin{table}[h]
    \centering
    \begin{tabular}{lcc}
        \toprule
        \rowcolor{gray!10} \textbf{Subset} & \textbf{Serialization Time (Hrs)} \\
        Commercial              & 23:50:57 \\
        NonCommercial           & 7:36:35 \\
        Other                   & 1:36:42 \\
      \midrule
        Total  & 33:04:14 \\
        \bottomrule
    \end{tabular}
    \caption{Total Serialization time  by PMC-OA subset}
    \label{tab:serialziation}
\end{table}

\subsection{Tokenized Caption Distribution}
\begin{figure*}[htbp]
    \centering
    \includegraphics[width=0.7\linewidth]{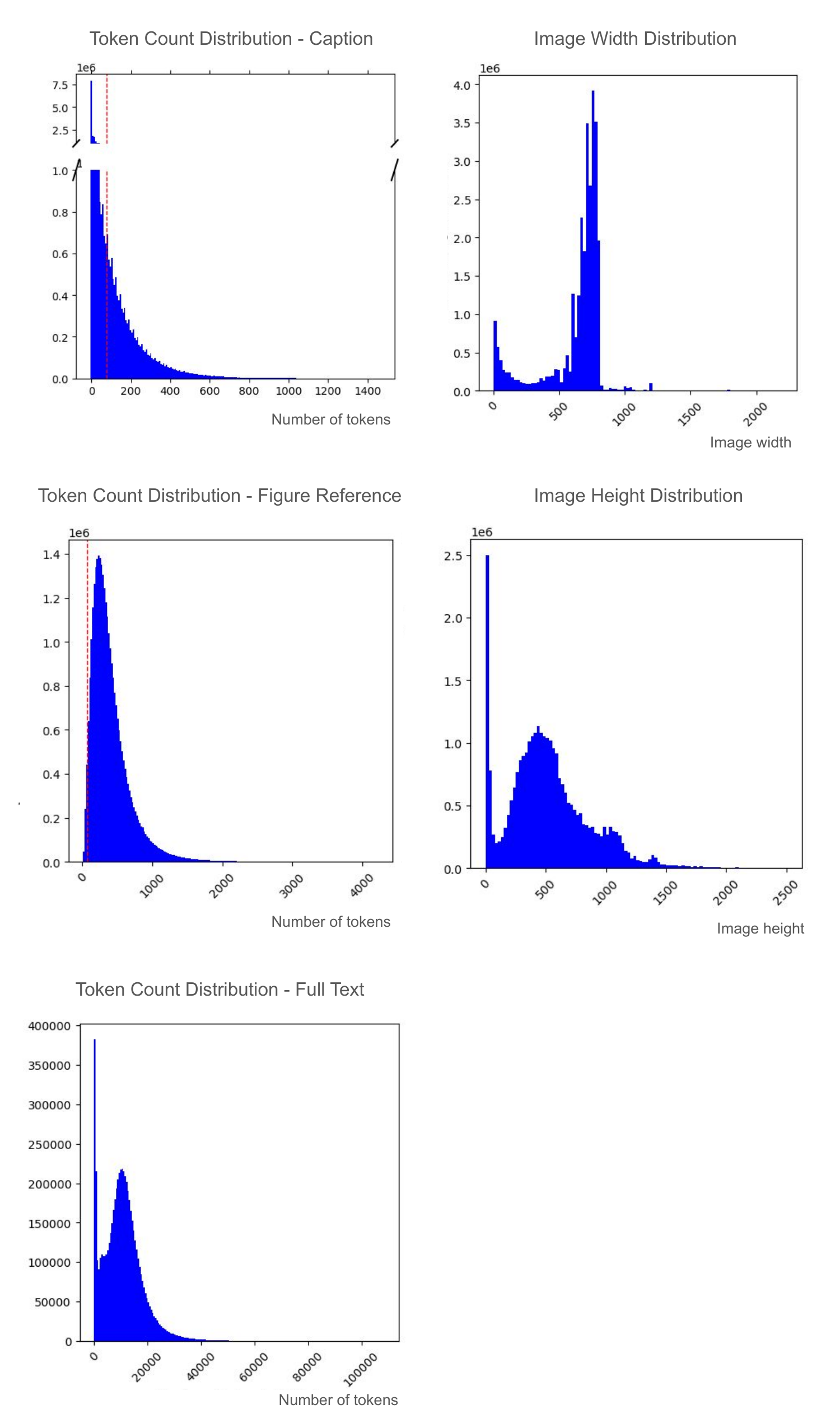}
    \caption{Distributions of token counts and image dimensions in the dataset. Histograms are shown for token counts in captions, figure references, and full text, as well as for image widths and heights. Outliers have been excluded to highlight the central tendencies and areas of higher data density.}
    \label{fig:histogram_token}
\end{figure*}

Note that the histogram in Figure~\ref{fig:histogram_token} shows a long right tail in the distribution of caption token lengths, with many captions exceeding the CLIP model's context length limit. \textbf{Need to provide numbers, mean min, max, median etc}

This issue is even more pronounced for figure reference.

\section{Concept Labeling: Additional Details}
\label{a:concept-labeler}

We developed an AI-assisted pipeline to categorize similar concepts within PMC-OA, as described in section Section \ref{cls_}. In summary, this process involves four stages: First, we define and organize similar images in clusters applying unsupervised clustering on image content, second, a group of 2 clinicians and 1 scientist use these clusters and taxonomies to create a hierarchical taxonomy for PMC-OA (see Figure \ref{fig:xml_markup_example}), then a group of  2 clinicians and 6 scientists use this taxonomy to annotate each cluster.  Lastly, metadata is propagated to each cluster instance. 

In this section, we provide additional details and experiments for each design choice.

\subsection{Dimensionality Reduction}

We used DINOv2 (ViT-L/14 distilled) to generate a 1024-dimensional vector for each image in PMC-OA . However, directly clustering such high-dimensional data can lead to poor performance due to the ``curse of dimensionality'', where the increased sparsity in high-dimensional spaces reduces the reliability of distance-based measures like those used in K-means clustering.

\begin{figure}
    \centering
    \includegraphics[width=0.9\linewidth]{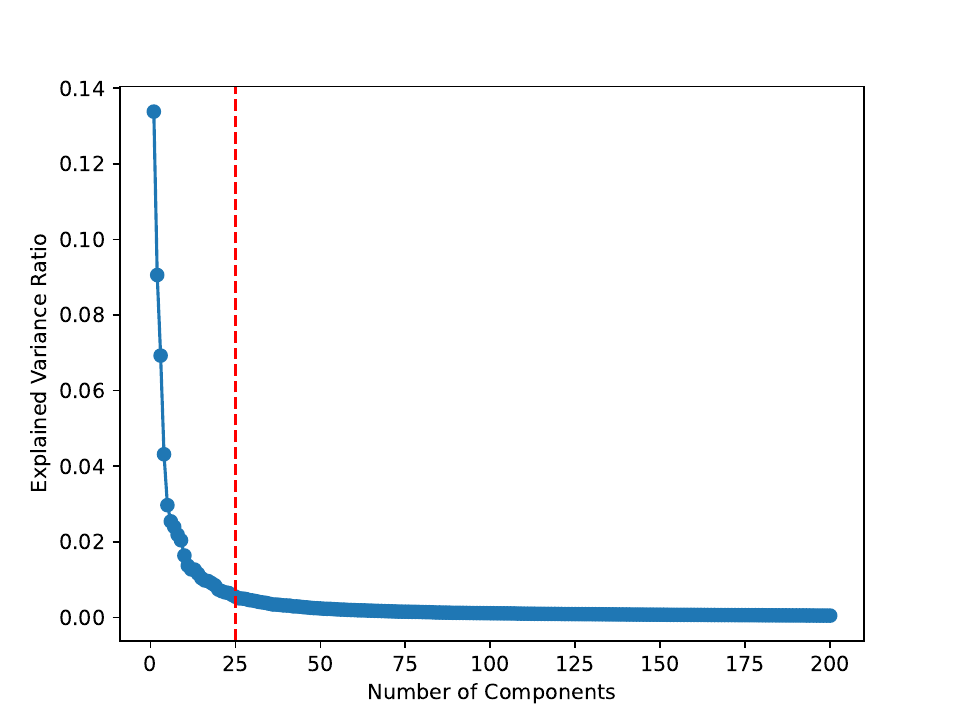}
    \caption{DinoV2 features Scree plot}
    \label{fig:DinoPCA}
\end{figure}

To address this, we applied PCA (Principal Component Analysis) to reduce the dimensionality of the embeddings. A scree plot analysis was performed to determine the minimum number of principal components required to retain 99\% of the data variance. As shown in Table \ref{fig:DinoPCA}, 25 principal components were sufficient to achieve this threshold. Consequently, we selected PCA (n=25) to transform the data before applying K-means clustering. 

\subsection{Over-Clustering}
We opt to over-cluster using KMeans with (K=2000), as this approach allows us to thoroughly explore the PMC-OA dataset. By focusing on fine-grained patterns and sub-groups, this strategy ensures a detailed analysis, particularly when the optimal number of clusters remains uncertain. This number was also selected to achieve an effective annotation time of at most  \textbf{48 hours per annotator}. 

\vspace{1em}

\noindent To do this at scale, we extract DINO-v2 features in parallel with NVIDIA Triton Inference. PCA and K-means are not parallelized. However, due to the simplicity of these approaches, they scaled to 24M pairs. Finally, cluster labels are stored along with image-uid. This is later used to propagate metadata to individual instances.

\subsection{Online Cluster Annotations Form}

Via a form hosted on Google Forms we asked two  practicing clinicians and 5 scientists to annotate each image cluster by answering the following questions:

\begin{enumerate}
\item \textit{Are the majority images within this cluster single panel or multiple panel?}
Options: 
\subitem Single Panels
\subitem Multiple panels with non-biomedical imaging
\subitem Multiple panels with biomedical imaging and plots
\subitem Multiple panels with biomedical imaging and assays

\item  \textit{Think of the main characteristics the MAJORITY of images in this cluster share in common. Given these characteristics, what is the most likely global class for this group?} Options:
\subitem Clinical Imaging
\subitem Microscopy
\subitem Immuno Assays
\subitem Illustrative Diagrams
\subitem Hand Drawn and Screen Based Visuals
\subitem Tables
\subitem Plots and Charts
\subitem PCR
\subitem Graph and Network
\subitem Scientific Formulae and Equations
\subitem Chemical Structures
\subitem Maps
\subitem Tools and Materials
\item \textit{Given these characteristics, what is the most likely local class for this group? (remember to use the hierarchical taxonomy provided)}
\end{enumerate}

\vspace{1em}

\noindent  Due to the quantity of high-resolution images, is not possible to compile all clusters together, thus 20 Google Forms, each containing 100 clusters were automatically created using Google Apps Script. Within each form, we provide the cluster image on top of each question to facilitate annotations.  Annotators are calibrated by doing a practice run on 20 examples (these annotations are not added to the analysis). Annotators were given two weeks \textbf{336 hours}) to fill all assigned forms. Annotators could only be assigned up to 1000/2000 clusters (10 forms). Lastly, forms were overlapped with a maximum of three annotators, meaning that annotations could have at most three labelers.

\subsection{Taxonomy Curators Statistics}

Table \ref{tab:cluster_annotators} shows statistics for taxonomy curators (denoted with $^*$). Curators have a minimum of 3 years of experience, with a median of 11 years and a maximum of 14 years. Both clinical curators have an additional PhD. All curators have wet-lab experience.

\subsection{Cluster Annotator Statistics}

Table \ref{tab:cluster_annotators} shows statistics for cluster annotators. Annotators have a minimum of 3 years of experience, with a median of 4 years and a maximum of 14 years. On average, annotators spent 17.66 hours to finish all assigned forms. Collectively, annotators spent 103 hours annotating all 2000 clusters.

\vspace{1em}

\noindent For our statistics, experience in the biomedical domain is measured from the time a person starts their graduate studies (such as a master's or Ph.D.) and continues to accumulate as they work in the field. For this work, undergraduate studies do not contribute to the experience count, meaning that time spent pursuing a Bachelor's degree or any other undergraduate-level studies is excluded from the total experience calculation, even when it's biomedical related.

\subsection{Dataset Taxonomy} 

The full delivered taxonomy is shown in Figure \ref{fig:xml_markup_example}. Examples of topics with they corresponding image cluster are shown in Table \ref{taxonomy-example}. In total the taxonomy spans 13 global topics enumerated below:

\begin{itemize}
   
    \item Clinical Imaging
    \item Microscopy
    \item Plots and Charts

    \item Immuno Assays

    \item Illustrative Diagrams
    \item Scientific Formulae and Equations
    \item Tables
    \item Hand Drawn and Screen Based Visuals
    \item Graph and Network
    \item Chemical Structures
    \item Maps
    \item Tools and Materials
    \item PCR (Polymerase Chain Reaction)
\end{itemize}

\noindent Notice that in contrast to the derived taxonomy, "Clinical and Scientific Image Data"  is not included as a topic. Instead we itemize this concept by its children (clinical imaging and microscopy). This design choice facilitates labeling. For the rest of this work,  clinical imaging and microscopy do not share any parent.

\subsection{Label Assignment and Propagation}
After labeling, each cluster has at most three expert-level annotations pertaining to panel type, global taxonomy, and local taxonomy.  To resolve annotations, we follow a three-step pipeline: 

\vspace{1em}

\noindent \textbf{1. Text Prepocessing} Since panel-type and global taxonomy are multiple-choice questions, no further reprocessing is necessary. Local taxonomy, however, consists of open questions (users are asked to follow the taxonomy as shown in Figure \ref{fig:xml_markup_example}). To this end, answers in this section are preprocessed by being lowercased, and white spaces and dashes are deleted.

\noindent  \textbf{2. Majorty vote} After preprocessing the final label for each field is resolved by taking the consensus of annotations by using the majority vote.

\noindent  \textbf{3. Label Propagation} The metadata for the labeled cluster is then propagated to each image within that cluster. This is accomplished by retrieving the cached cluster label and image-uid (see  Over-Clustering section) along with the corresponding resolved clustered metadata and adding this information to each item in within the serliaized data.

\begin{table}[hbt!]
\small
\centering
\begin{adjustbox}{max width=\textwidth}
\begin{tabular}{lccccc}
\toprule
\rowcolor{gray!10} \textbf{Concept}          & \textbf{Min} & \textbf{Max}         & \textbf{Mean} & \textbf{Median}  &  \textbf{IQR}  \\
Panel           & 0.0           &   66.66           & 13.54            &   0.0       & 33.33    \\
Global         & 0.0           &   66.66           & 8.39         &   0.0       & 0.0    \\
\hline
Local              & 0.0           &   75.0           & 19.91        &   0.0       & 50.00    \\

\bottomrule
\end{tabular}
\end{adjustbox}
\caption{Statistical overview of inter-annotator disagreement (lower is better).}
\label{tab:interanotator-statistics}
\end{table}

\begin{figure*}[hbt!]
    \centering
    \includegraphics[width=1\linewidth]{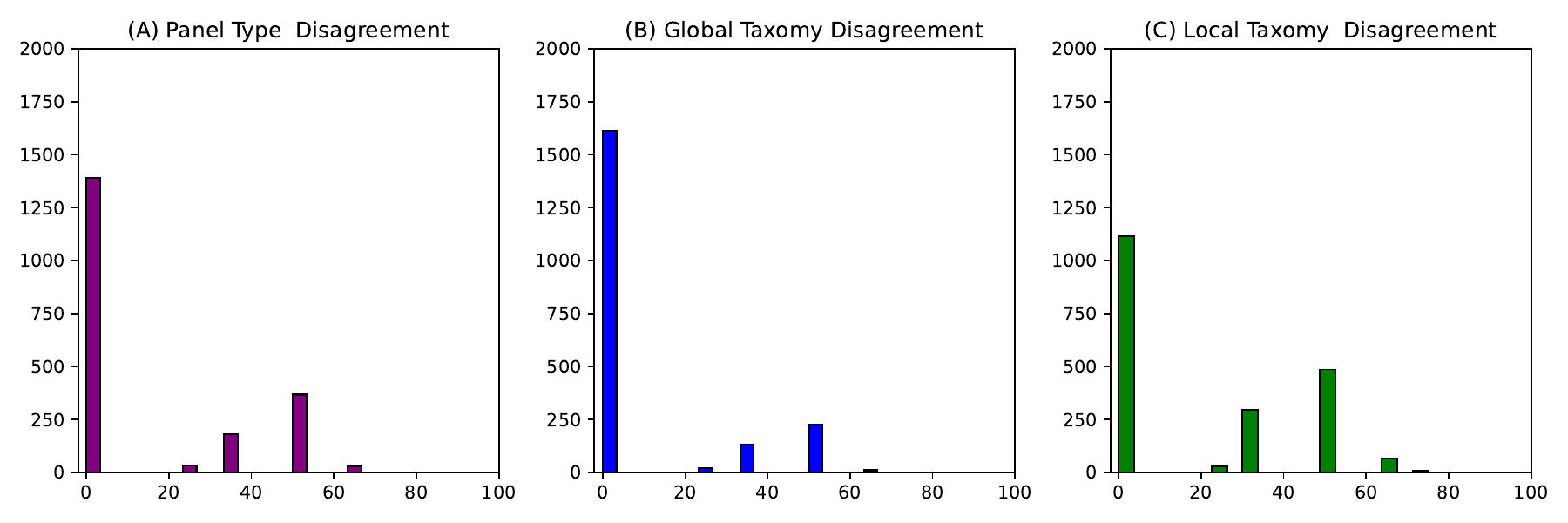}
    \caption{Inter-annotator disagreement (lower is better).}
    \label{fig:dis}
\end{figure*}

\subsection{Inter-annotator Disagreement}

Table \ref{tab:interanotator-statistics} shows a statistical overview of inter-annotator disagreement. The median disagreement for all concepts is 0.0, while the maximum disagreement (present in local taxonomy) is 19.91\%, highlighting a high inter-annotator agreement greater than 80\% for all concepts. 'Easier' concepts, such as global taxonomy, show a low mean disagreement of 8.39\%. Figure \ref{fig:dis} shows a histogram summarizing these findings.

\section{Data Upload}

Data is uploaded to HuggingFace using "upload\_large\_folder"  with 40 workers. This upload mechanism enables parallelism but relies on an already structured dataset.

\begin{verbatim}

from huggingface_hub import HfApi

api = HfApi(token=HF_TOKEN)
api.upload_large_folder(
    repo_type    = "dataset",
    repo_id      = REPO_ID,
    folder_path  = LOCAL_PATH,
    num_workers  = 40)
\end{verbatim}



\section{Model Training}
\label{a:model-training}

\subsection{Base Model Selection}
\label{a:base-model}
In biomedical model development, it is common to skip ablations when selecting a strong base model for continual pretraining, often defaulting to models that yield state-of-the-art results in general domains. To this end, we use the validation sets of four random datasets to identify a robust base model for further fine-tuning (see Table \ref{tab:_base_model_performance}). Our analysis reveals that, generally, ViT models trained on datasets such as CommonPool, Liabo2B, and WebLi achieve the strongest performance in the selected biomedical domains, while previously favored models like CoCa show the weakest results. Based on these findings, we select CLIP, ViT-L-14 Base as our model for subsequent experiments.

\subsection{Modeling Hyperparameters}
We detail the key configurations for training our model as follows:

\begin{itemize}
    \item \textbf{Batch Size and Accumulation}: We use a batch size of 1024 per GPU on 4 GPUs with a batch accumulation frequency of 2, yielding an effective batch size of 8192.
    \item \textbf{Learning Rate}: We perform a sweep of learning rates from $1e-6$ to $1e-8$. We select the biggest learning rate with stable training curve. To this end ,
    We use a learning rate of 1e-6 and a warmup phase of 1000 steps.
    \item \textbf{Optimizer}: 
    We use the Adam optimizer with parameters set to $\beta_{1}=0.9$ and $\beta_{2}=0.95$, and a weight decay of 0.2.
    \item \textbf{Floating-Point Precision}: All computations are performed with 32-bit floating-point precision.
\end{itemize}

\noindent Table  \ref{tab:hyperparam_search_grid} summarizes all parameters used during training.

\begin{table}[h]
    \centering
    \begin{tabular}{llr}
       \toprule
       \rowcolor{gray!10}  \textbf{Model} & \textbf{Pretraining Dataset} & \textbf{Mean} \\
        ViT-L-14               & commonpool      & 38.465 \\
        ViT-B-32               & laion2b                       & 37.892 \\
        ViT-SO400M-14-SigLIP   & webli                              & 35.010 \\
        ViT-L-14               & laion2b                 & 33.951 \\
        convnext-larged-320   & laion2b         & 33.773 \\
        ViT-L-14               & datacomp            & 33.148 \\
        EVA02-B-16             & merged2b s8b b131k                 & 32.995 \\
        RN50-quickgelu         & openai                             & 32.736 \\
        ViT-B-16-SigLIP-384    & webli                              & 32.482 \\
        ViT-B-16-SigLIP        & webli                              & 32.399 \\
        ViT-B-16-SigLIP-256    & webli                              & 31.762 \\
        EVA02-L-14             & merged2b s4b b131k                 & 31.680 \\
        ViT-B-16-SigLIP-512    & webli                              & 31.319 \\
        ViT-B-32               & commonpool        & 30.614 \\
        coca-ViT-B-32          & laion2b  & 29.133 \\
        ViT-B-32               & datacomp              & 29.077 \\
        convnext-B        & laion2b          & 28.701 \\
        ViT-B-16               & laion2b                & 25.786 \\
        coca-ViT-L-14          & laion2b                 & 25.212 \\
        \bottomrule
    \end{tabular}
    \caption{Average Accuracy of  Base Models (with Corresponding Pretraining Datasets) on  5 biomedical classification tasks}
    \label{tab:_base_model_performance}
\end{table}

\begin{table*}[h!]
\centering

\begin{tabular}{llll} 
\toprule
\rowcolor{gray!10} \multicolumn{1}{c}{\textbf{Model}} & \textbf{Continual Training Data} & \textbf{Hyperparameters} & \textbf{Values} \\ 
\midrule
\multirow{18}{*}{\textbf{CLIP: ViT-L-14 (commonpool)}} & Full Data (24M) & batch size (per GPU)        & 1024 \\ 
 & & GPUs            & 4xH100 \\
 & & accumulation frequency             & 2   \\ 
  & & effective batch size            & 8192 \\
 & & learning rate               & $1\mathrm{e}{-6}$   \\ 
 & & beta1                       & 0.9 \\ 
 & & beta2                       & 0.95 \\ 
 & & warmup                      & 1000 \\ 
 & & epochs                      & 9 \\ 
 & & precision                   & FP32 \\ 
 & & gradient clipping norm      & 1.0 \\ 
 & & dataset type                & WebDataset \\ 
\cmidrule{2-4} 
 & Concept Balanced Data (8M) & batch size (per GPU)        & 1024 \\ 
  & & GPUs            & 4xH100 \\
 & & accumulation frequency             & 2   \\ 
  & & effective batch size            & 8192 \\
 & & learning rate               & $1\mathrm{e}{-6}$   \\ 
 & & beta1                       & 0.9 \\ 
 & & beta2                       & 0.95 \\ 
 & & warmup                      & 1000 \\ 
 & & epochs                      & 27 \\ 
 & & precision                   & FP32 \\ 
 & & gradient clipping norm      & 1.0 \\ 
 & & dataset type                & WebDataset \\ 
\cmidrule{2-4} 
 & Concept Filtered Data (6M) & batch size (per GPU)        & 1024 \\  
  & & GPUs            & 4xH100 \\
 & & accumulation frequency             & 2 \\
 & & effective batch size            & 8192 \\
 
 & & learning rate               & $1\mathrm{e}{-6}$  \\ 
 & & beta1                       & 0.9 \\ 
 & & beta2                       & 0.95 \\ 
 & & warmup                      & 1000 \\ 
 & & epochs                      & 36 \\ 
 & & precision                   & FP32 \\ 
 & & gradient clipping norm      & 1.0 \\ 
 & & dataset type                & WebDataset \\ 
 \midrule

\end{tabular}
\caption{Hyper-parameters used for continual pretraining.}
\label{tab:hyperparam_search_grid}
\end{table*}
\newpage


\begin{table*}[t]
\centering
\caption{Provenance of evaluation benchmark. Each row lists dataset name (with it's corresponding citation),  task description, image modality, and number of classes in each corresponding tasks.}
\begin{adjustbox}{max width=\textwidth}
\begin{tabular}{p{4.5cm}p{4cm}p{4.5cm}p{2cm}}
\toprule
 \textbf{Dataset} & \textbf{Task} & \textbf{Modality} & N. Classes \\ \midrule

\rowcolor{gray!10}\textbf{Cell Biology} &&& \\

\multicolumn{3}{l}{\textit{Cell Cycle \& Stage Identification}} \\
BBBC048 (BF) \cite{eulenberg2017reconstructing} & Cell cycle phase & Fluresence Microscopy & 7  \\
BBBC048 (DF) \cite{eulenberg2017reconstructing} & Cell cycle phase & Fluresence Microscopy  & 7\\
BBBC048 (EF) \cite{eulenberg2017reconstructing} & Cell cycle phase & Fluresence Microscopy & 5 \\

\multicolumn{3}{l}{\textit{Cell Profiling}} \\
PCST Contour \cite{Burgess2024-lg} & Cell contour & Fluresence Microscopy & 3 \\
PCST-Texture \cite{Burgess2024-lg} & Cell texture & Fluresence Microscopy  & 3\\
PCST Eccentricity \cite{Burgess2024-lg} & Cell eccentricity & Fluresence Microscopy & 3\\

 \multicolumn{3}{l}{\textit{Cell \&  Structure Identification}} \\
Fluorescence Cells \cite{lozano2024mu} & Organisms and structures & Fluresence Microscopy & 13  \\
EMPIAR SBF-SEM \cite{iudin2023empiar} & Organisms and structures & Electron Microscopy  & 5  \\
ICPR2020 Pollen \cite{Battiato2020POLLEN13KAL} & Pollen structures & Fluresence Microscopy  & 4 \\


\midrule
\rowcolor{gray!10} \textbf{Pathology} &&&\\
\multicolumn{3}{l}{\textit{Cytology}} \\

Acevedo et al 2020 \cite{Acevedo2020-ja} & White blood cell & Light Microscopy  (Giemsa) & 8 \\
Jung et al 2022 \cite{jung2022wbc} & White blood cell & Light Microscopy (Giemsa)  &  5\\ 
Pap Smear 2019 \cite{Hussain2020-vq} & Pap smear grading & Light Microscopy  (Pap Smear) & 4\\ 

\multicolumn{3}{l}{\textit{Neoplastic Histopathology}} \\

Kather et al 2016 \cite{Kather2016-db} & Colorectal tissue & Light Microscopy  (H\&E) & 8\\
LC25000 (Lung)  \cite{borkowski1912lung} & Lung tissue classification & Light Microscopy  (H\&E) & 3 \\
 PCAM \cite{litjens20181399}  & Lymph node classification &  Light Microscopy  (H\&E) & 2\\

LC25000 (Colon) \cite{borkowski1912lung} & Colon tissue classification & Light Microscopy  (H\&E) & 2 \\  

\multicolumn{3}{l}{\textit{Non-neoplastic Histopathology}} \\
Tang et al 2019 \cite{Tang2019-fy} & Amyloid morphology & Light Microscopy  (IHC) & 4\\
Wong et al 2022 \cite{Wong2022-nm} & Amyloid morphology & Light Microscopy  (IHC) & 4\\
Nirschl et al 2018 \cite{Nirschl2018-pc} & Clinical chronic heart failure & Light Microscopy  (H\&E) & 2 \\
Wu et al 2023 \cite{Wu2023-ht} & Mitochondrial morphology & Light Microscopy  (IHC) & 2\\ 

\midrule
\rowcolor{gray!10} \textbf{General Microscopy} &&& \\ 

Micro-Bench Submodality \cite{lozano2024mu} & Microscopy Submodality &  Microscopy & 6  \\

Micro-Bench Stain \cite{lozano2024mu} & Microscopy Stain &  Microscopy & 6  \\
Micro-Bench Domain \cite{lozano2024mu} & Microscopy Domain &  Microscopy & 6  \\
Micro-Bench Modality \cite{lozano2024mu} & Microscopy Modality &  Microscopy & 3  \\
\midrule

\rowcolor{gray!10} \textbf{Radiology} &&& \\ 
\multicolumn{3}{l}{\textit{Diagnostics}} \\
Chexpert \cite {irvin2019chexpert}  &Chest X-Rays Findings & Chest X-Ray  & 5 \\ 
RSNA 2018 & Chest X-Rays Findings & Chest X-Ray  & 2\\ 
BreastMNIST  \cite{al2020dataset} & Breast cancer diagnosis   & Breast Ultrasound & 2 \\ 
\midrule

\rowcolor{gray!10} \textbf{Dermatology}  &&&\\
HAM1000  \cite{DVN/DBW86T_2018} & Common  skin lesions& Dermatoscope &  7 \\ 
\midrule
\rowcolor{gray!10} \textbf{Surgery} &&& \\
Dresden  Anatomy Dataset \cite{carstens2023dresden} & Anatomy in surgery  & Laparoscopic
Surgery &  7x2  \\ 
\midrule
\rowcolor{gray!10} \textbf{Ophthalmology} &&&\\
DeepDRiD \cite{liu2022deepdrid} & Diabetic retinopathy  & Retina Fundus Images & 4\\
\bottomrule
\label{table:benchmark_sources}
\end{tabular}
\end{adjustbox}
\end{table*}

\



\begin{table*}[h!]
\small
\centering
\begin{adjustbox}{max width=\textwidth}
\begin{tabular}{lccccc}
\toprule
\rowcolor{gray!10} \textbf{Text Statistics}          & \textbf{Min} & \textbf{Max}         & \textbf{Median} & \textbf{IQR} & \textbf{Total}        \\
Caption Token Length              & 4            & 1324               & 23             &   63      & $ 4.5 \times 10^5$    \\
Caption Character Length          & 10            & 3287                & 78  & 149 & $1.10\times 10^6$ \\
\hline
\rowcolor{gray!10} \textbf{Image Statistics}         &              &                      &                 &              &                       \\
Image Width (pixels)              &  70           & 1024               & 1024             & 1          &                       \\
Image Height (pixels)             &  70           & 1024                & 768             & 87          &                       \\
Image Area (pixels²)              & 4900            & 1048576 & 786432         & 109206     &                       \\ 

\bottomrule
\end{tabular}
\end{adjustbox}
\caption{Overview of  Biomedical Flickr statistics, detailing text token and character lengths, and image dimensions. Tokens are generated using the BPE  tokenizer from the \texttt{tiktoken} library}
\label{tab:flickr-statistics}
\end{table*}


\section{Evaluation}

\subsection{Closed VQA  Benchmark}
\subsubsection{Closed VQA Formulation}
A total of 39 existing classification tasks are formulated as multiple-choice visual question answering. The following subsection provides additional details for evaluation re-formulation.

 We first collect the test set of each dataset, yielding image-label pairs.
$$ D_i = \{(x_{i}^1, l_{i}^1), \dots, (x_{i}^{N_i}, l_{i}^{N_i})\} $$

\noindent where \( N_i \) corresponds to the total number of samples in the test subset of the i-th dataset. 

For each image-label pair, the label   \( l_{i}^j \) corresponds to one of  \( M_i \) possible classes defined for the i-th dataset..
\[
l_{i}^j \in C_i,
\]  
where \( C_i = \{c_1, c_2, \dots, c_{M_i}\} \) , $|C_i|=M_i $

\vspace{1em}

\noindent To convert each classification task into a closed VQA task, we define a mapping function \( f_i \) for each  dataset. This function maps a given label \( l_{i}^j \) to a human written  textual description:

\[
f_i: l_{i}^j \to t_{i}^j,
\]

\noindent where \( t_{i}^j \) is the textual descriptor of label \( l_{i}^j \). This process is applied to the entire dataset:

$$ D_i = \{(x_{i}^1, t_{i}^1), \dots, (x_{i}^{N_i}, t_{i}^{N_i})\} $$

\noindent Then the reminder of (incorrect) classes textual descriptions are added to each data point to create a multiple-choice list: \( A_i = \{t_{i}^j , a_i^1, a_i^2, \dots, a_{M_i^{i-1}}\} \). Lastly, a random permutation is applied, storing the position of the correct label after this operation $k_{i}^j$.

These operations convert the initial dataset to a collection of image-text pairs, where each image \( x_i \) is associated with:  
1. A list of multiple-choice answers 
2. The correct index of the label within this list, denoted as \( k_i \):  

\[
D_i = \{(x_{i}^1, A_{i}^1, k_{i}^1), \dots, (x_{i}^{N_i}, A_{i}^{N_i}, k_{i}^{N_i})\}.
\]

\noindent Where \( \pi \) denotes the random permutation function applied to the answers in \( A_i^{\pi(j)}\).

\subsubsection{ Closed VQA Evaluation}

\noindent
All evaluated contrastive models have a vision encoder $E_{image}$ and text encoder $E_{text}$. We first compute the image embedding, $z_{x_i}=E_{image}(\textbf{x}_i^j)$, along with  each candidate answer: $z_{a_{i}^\pi(j)}=E_{text}(a_{i}^\pi(j) )$ for $l\in[1,M_i]$. Then we compute the cosine similarity score for each caption, $s_{ij}=z_{a_{i}^\pi(j)}\cdot z_{x_i}^T$ for $l \in[1,M_i]$. The option with the largest $s_{ij}$ is then assigned as the final prediction. If argmax($s_{ij}$) has the same index as the corresponding correct answer $k_{i}^j$ the question is marked as correct, incorrect otherwise.

\subsubsection{Closed VQA Conversion Prompts}

If a dataset did not explicitly contain a Closed-VQA form, then a group consisting of a biomedical informatics, pathologist converted each class to unique its corresponding caption. 

\subsection{Retrieval Benchmark Evaluation}

Given a dataset of images and captions
\[
D_c = \{(x^1, c^1), \dots, (x^{N_i}, c^{N_i})\}.
\]

\noindent We evaluate retrieval performance using  Recall@k, using the following protocol:

All evaluated contrastive models have a vision encoder $E_{image}$ and text encoder $E_{text}$. We first compute the image embedding, $z_{x_i}=E_{image}(\textbf{x}^i)$, along with  each caption: $z_{c_i}=E_{text}( c^i )$ for $l\in[1,M_i]$. Then we compute the cosine similarity score for each caption, $s_{ij}=z_{c^i}\cdot z_{x_i}^T$ for $l \in[1,N_i]$. Captions are arranged from the largest to smallest similarity ($s_{ij}$). If the correct caption is within the first $k$-th arranged items, then the option is considered relevant, irrelevant otherwise. lastly, we calculate Recall@k using the following equation:

\[
\text{Recall@k} = \frac{\text{Number of relevant items in the top } k \text{ results}}{\text{Total number of relevant items in the dataset}}
\]

\subsection{Computing Confidence Intervals}
\label{sec:cofidence_intervals} 
Error bars represent 95\% confidence intervals (CI) computed via nonparametric bootstrapping using the SciPy $stats\text{.}bootstrap$ function with 1000 resampling and default settings.

\section{Flickr Dataset Description} \label{a:flickr_data}
The Biomedical Flickr dataset consists of 7k image-caption pairs retrieved from flicker channels with permissive licenses.  It mostly spans microscopy. Table \ref{tab:flickr-statistics} shows statistics for the dataset. Table \ref{flickr-example} shows 10 random samples from the dataset.

\section{Compute Environment}
\label{a:compute}
Experiments are performed in a local on-prem university compute environment using 24 Intel Xeon 2.70GHz CPU cores, 8 Nvidia H100 GPUs, 16 Nvidia A6000 GPUs, and  40 TB of Storage.

\newpage

\begin{table*}[h!]
\centering
\begin{tabular}{l r}
\rowcolor{gray!10}\textbf{MeSH Term} & \textbf{Frequency} \\ \hline
Humans & 2,189,713 \\
Female & 990,873 \\
Male & 897,332 \\
Animals & 775,314 \\
Adult & 521,585 \\
Middle Aged & 483,806 \\
None & 375,532 \\
Aged & 371,170 \\
Mice & 288,320 \\
Young Adult & 205,580 \\
Adolescent & 200,700 \\
Retrospective Studies & 178,740 \\
Child & 166,187 \\
COVID-19 & 149,167 \\
Cross-Sectional Studies & 135,551 \\
Risk Factors & 135,111 \\
Treatment Outcome & 131,638 \\
Aged, 80 and over & 125,351 \\
Signal Transduction & 109,685 \\
SARS-CoV-2 & 100,287 \\
Cell Line, Tumor & 100,154 \\
Surveys and Questionnaires & 98,150 \\
Prospective Studies & 96,729 \\
Prognosis & 89,765 \\
Rats & 89,433 \\
Pregnancy & 88,439 \\
Child, Preschool & 79,751 \\
Mutation & 79,073 \\
Biomarkers & 77,093 \\
Disease Models, Animal & 76,046 \\
Cell Proliferation & 75,080 \\
Time Factors & 75,034 \\
Mice, Inbred C57BL & 72,500 \\
Infant & 71,594 \\
Pandemics & 70,915 \\
China & 68,963 \\
Algorithms & 64,904 \\
Neoplasms & 64,674 \\
Cohort Studies & 64,563 \\
Reproducibility of Results & 62,995 \\
Phylogeny & 62,412 \\
Prevalence & 61,952 \\
Apoptosis & 61,057 \\
Cells, Cultured & 58,811 \\
Cell Line & 57,933 \\
Gene Expression Profiling & 57,680 \\
Brain & 57,472 \\
Case-Control Studies & 57,044 \\
Quality of Life & 56,170 \\
Infant, Newborn & 55,529 \\ \hline
\end{tabular}
\caption{Top 50 most common MeSH Terms and their frequencies}
\label{tab:mesh_terms}
\end{table*}


\label{a:Dataset_Taxonomy}

\begin{table*}[hbt!]
\small
\centering
\begin{adjustbox}{max width=\textwidth}
\begin{tabular}{lclll}
\toprule
\rowcolor{gray!10} & \textbf{Years of experience} &	\textbf{Field of study} & \textbf{Role} \\
Annotator 1 & 3 & Developmental biology & PhD Student \\
Annotator 2 & 3 & Microbiology & PhD Student \\
Annotator 3$^*$ & 3 & Biomedical Informatics & PhD Student \\

Annotator 4 & 5 & Biomedical Informatics & PhD Student \\
Annotator 5 & 3 & Genetics & PhD Student \\
Annotator 6 & 9 & Biomedical Informatics & MD-PhD Student \\
Annotator 7$^*$ & 11 & Biomedical Engineering, Molecular Biology, Surgical Data Science, ML & Industry Director, Post-doc \\
Annotator 8$^*$ & 14 & Pathology, Cell biology, Neuroscience & Attending Pathologist, Post-doc \\

\bottomrule
\end{tabular}
\end{adjustbox}
\caption{Description of cluster annotators. Years of experience include years of research or laboratory experience in a biology/biomedical or microscopy related discipline. $^*$ Annotator developed taxonomy. }
\label{tab:cluster_annotators}
\end{table*}

\begin{table*}[]
\centering
    \begin{tabular}{lcccc}
    \toprule
\rowcolor{gray!10}    \textbf{Category} & \textbf{Other} & \textbf{Commercial}& \textbf{Noncommercial} & \textbf{Total} \\
Scientific Formulae and Equations & 2,322 & 20,384 & 6,182 & 28,888 \\
PCR & 2,307 & 25,353 & 7,693 & 35,353 \\
Tools and Materials & 10,320 & 210,740 & 51,339 & 272,399 \\
Maps & 18,700 & 264,092 & 41,473 & 324,265 \\
Hand Drawn and Screen Based Visuals & 24,690 & 356,047 & 76,404 & 457,141 \\
Graph and Network & 26,453 & 415,737 & 83,470 & 525,660 \\
Tables & 26,190 & 269,384 & 343,455 & 639,029 \\
Immuno Assays & 38,055 & 651,339 & 215,238 & 904,632 \\
Chemical Structures & 92,881 & 839,082 & 196,119 & 1,128,082 \\
Clinical Imaging & 82,349 & 1,078,901 & 766,908 & 1,928,158 \\
Microscopy & 101,617 & 1,818,302 & 597,413 & 2,517,332 \\
Illustrative Diagrams & 140,925 & 2,227,690 & 551,380 & 2,919,995 \\
Plots and Charts & 542,718 & 9,426,147 & 2,394,358 & 12,389,066 \\ 
    \bottomrule
    \end{tabular}
    \caption{Number of images by global taxonomy concepts.} 
    \label{tab:global_taxonomy}
\end{table*}

\begin{table*}[]
\centering
\begin{tabular}{lccccc}
\toprule
\rowcolor{gray!10} \textbf{License Type}       & \textbf{Number of Articles} \\
CC0                               & 132592                \\
CC BY                             & 3795419               \\
CC BY-SA                          & 1287                  \\
CC BY-ND                          & 7900                  \\
CC BY-NC                          & 771755                \\
CC BY-NC-SA                       & 275638                \\
CC BY-NC-ND                       & 642982                \\
Other                             & 414857                \\
\bottomrule
\end{tabular}
\caption{Number of articles by license type. As described in the PMC website: "Commercial use allowed: CC0, CC BY, CC BY-SA, CC BY-ND. Non-commercial use only: CC BY-NC, CC BY-NC-SA, CC BY-NC-ND. Other: no machine-readable Creative Commons license, no license tagged, or a custom license."}
\label{tab:license-statistics}
\end{table*}

\begin{table*}[t]
\centering
\small
\begin{adjustbox}{max width=\textwidth}
\begin{tabular}{p{10cm}p{3cm}p{3cm}p{2cm}}
\toprule
\rowcolor{gray!10} \textbf{Cluster Example} & \textbf{Global Taxonomy} & \textbf{Local Taxonomy} & \textbf{Multi-panel} \\
\includegraphics[width=10cm, height=1cm]{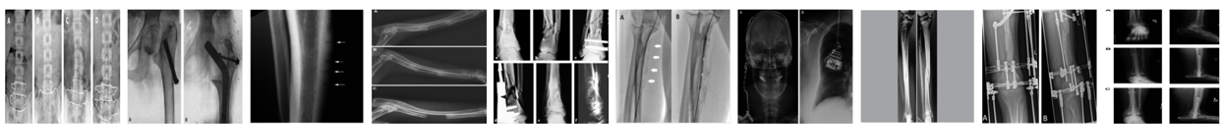} & \vspace{-2em} Clinical Imaging Data & \vspace{-2em} x-ray radiography & 
 \vspace{-2em} \checkmark \\

\includegraphics[width=10cm, height=1cm]{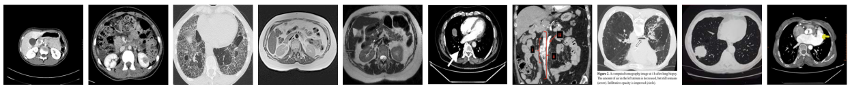} & \vspace{-2em} Clinical Imaging Data & \vspace{-2em}  computerized tomography & \vspace{-2em} \checkmark \\

\includegraphics[width=10cm, height=1cm]{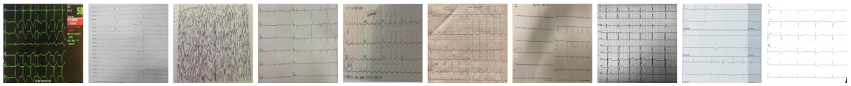} &\vspace{-2em}  Clinical Imaging Data & \vspace{-2em} electrocardiogram& \vspace{-2em} \textbf{\texttimes}  \\

\includegraphics[width=10cm, height=1cm]{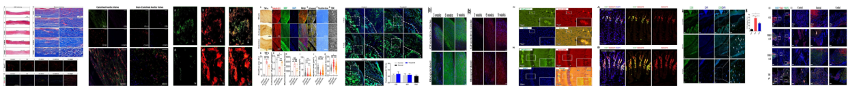} & \vspace{-2em} Microscopy & \vspace{-2em} fluorescence microscopy & \vspace{-2em} \checkmark \\

\includegraphics[width=10cm, height=1cm]{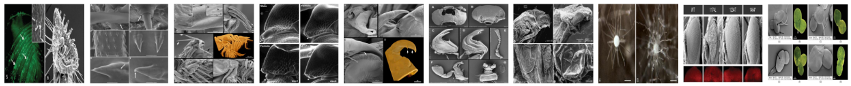} & \vspace{-2em} Microscopy & \vspace{-2em} electron microscopy &\vspace{-2em}  \checkmark \\
\includegraphics[width=10cm, height=1cm]{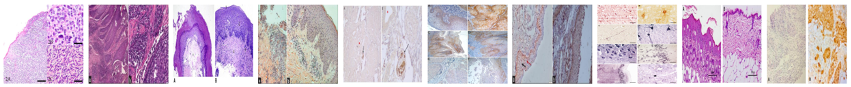} & \vspace{-2em} Microscopy &\vspace{-2em}  light microscopy & \vspace{-2em} \checkmark \\

\includegraphics[width=10cm, height=1cm]{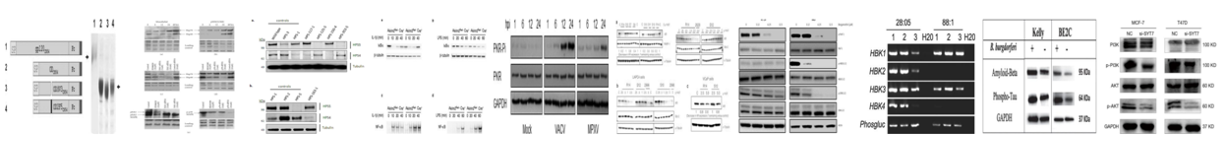} &\vspace{-2em} Immuno Assays & \vspace{-2em} gel electrophoresis & \vspace{-2em} \checkmark \\

\includegraphics[width=10cm, height=1cm]{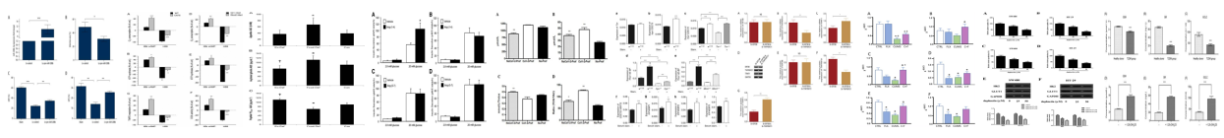} & \vspace{-2em} Plots and Charts &\vspace{-2em}  bar plot & \vspace{-2em} \checkmark \\

\includegraphics[width=10cm, height=1cm]{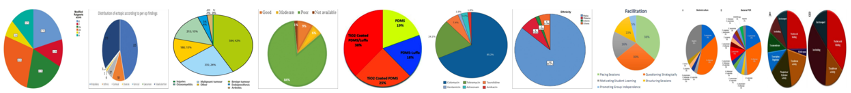} &\vspace{-2em}  Plots and Charts &\vspace{-2em}  pie chart & \vspace{-2em} \textbf{\texttimes} \\

\includegraphics[width=10cm, height=1cm]{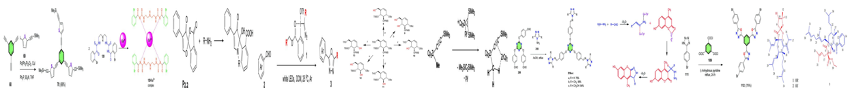} & \vspace{-2em}  Chemical Structures & \vspace{-2em} 2D chemical reaction  & \vspace{-2em} \textbf{\texttimes} \\

\includegraphics[width=10cm, height=1cm]{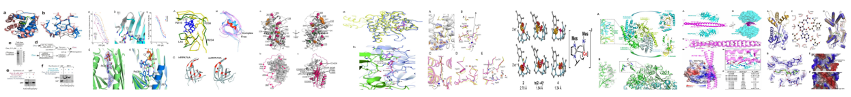} &\vspace{-2em}  Chemical Structures & \vspace{-2em} 3D chemical structure & \vspace{-2em} \textbf{\texttimes} \\

\includegraphics[width=10cm, height=1cm]{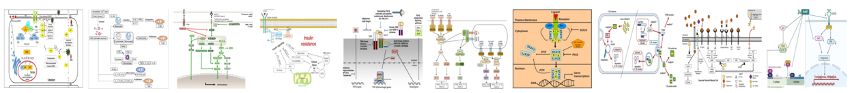} & \vspace{-2em} Illustrative Diagrams & \vspace{-2em}  signaling pathway & \vspace{-2em} \textbf{\texttimes} \\
\includegraphics[width=10cm, height=1cm]{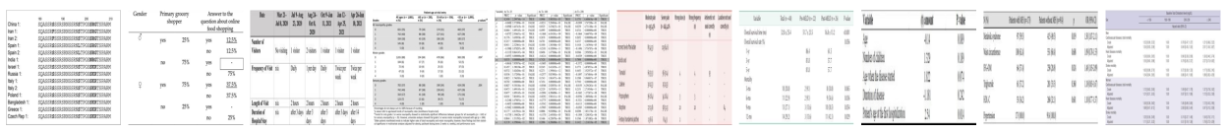} & \vspace{-2em} Tables & \vspace{-2em}  table & \vspace{-2em} \textbf{\texttimes} \\

\includegraphics[width=10cm, height=1cm]{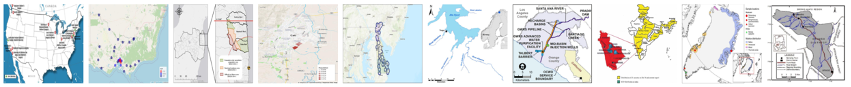} & \vspace{-2em} Maps & \vspace{-2em}  map &\vspace{-2em}  \textbf{\texttimes} \\

\includegraphics[width=10cm, height=1cm]{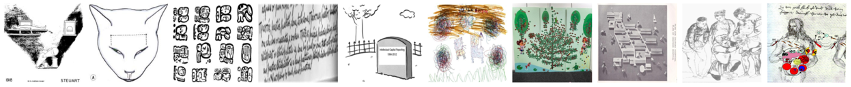} & \vspace{-2em} Drawings & \vspace{-2em}  drawing & \vspace{-2em} \textbf{\texttimes} \\

\includegraphics[width=10cm, height=1cm]{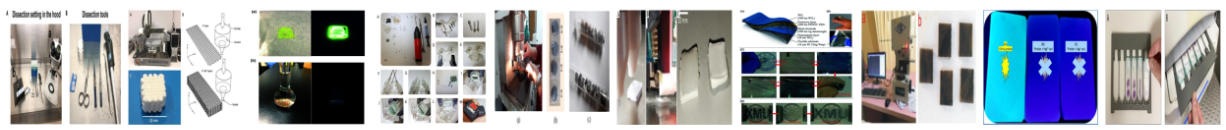} & \vspace{-2em} Tools and Materials & \vspace{-2em} lab equipment & \vspace{-2em} \checkmark  \\

\includegraphics[width=10cm, height=1cm]{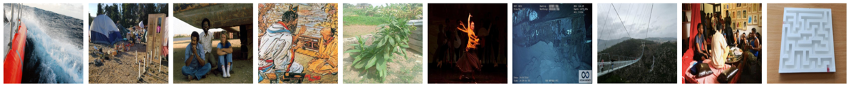} & \vspace{-2em} Natural Images & \vspace{-2em} natural image & \vspace{-2em}  \textbf{\texttimes}  \\
\bottomrule
\end{tabular}
\end{adjustbox}
\caption{Taxonomy of clusters with example images. Images resized to a uniform width of 10cm and height of 1cm. Column widths adjusted to fit the page.}
\label{taxonomy-example}
\end{table*}

\begin{table*}[t]
\centering
\small
\begin{adjustbox}{max width=\textwidth}
\begin{tabular}{p{2cm}p{18.5cm}}
\toprule
\rowcolor{gray!10} \textbf{Image} & \textbf{Caption} \\

\includegraphics[width=2.3cm, height=2.3cm]{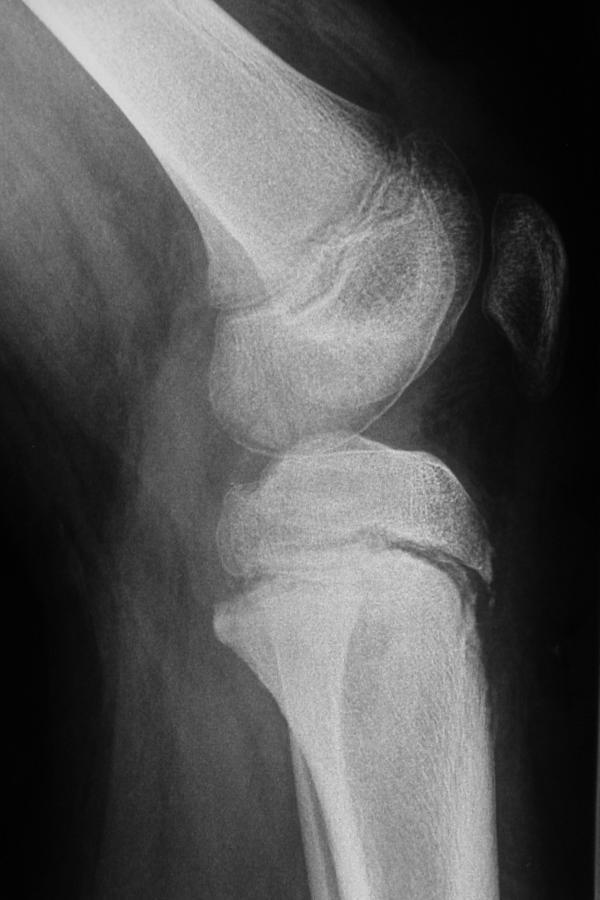}  &   \vspace{-7.2em}  Four weeks after the accident: the radiograph shows good alignment (lateral view). \\

\includegraphics[width=2.3cm, height=2.3cm]{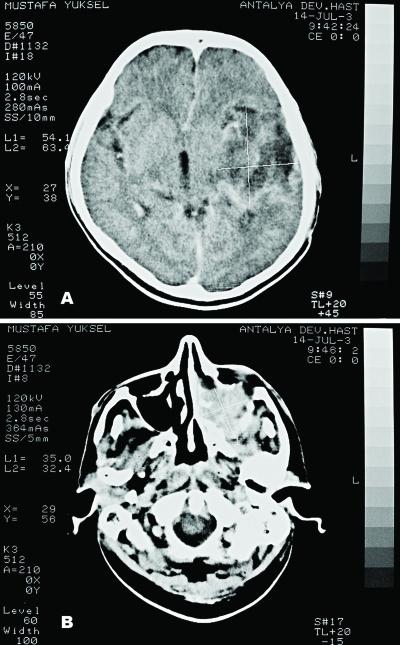}  &   \vspace{-7.2em}  CT-images belong to brain and maxillary sinus. Ct-images taken before any procedure applied illustrating two separate tumors in the brain (2a) and left-maxillary sinus (2b). Note the enhancing heterogeneous tumor at the left-temporoparietal lobe shifting the midline to the right (2a); and invasion of the tumor (T4) in the left-maxillary sinus into the adjacent tissues (2b).\\

\includegraphics[width=2.3cm, height=2.3cm]{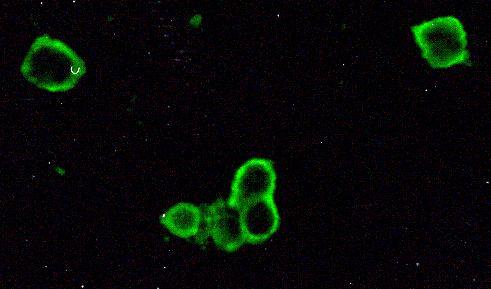}  &  \vspace{-7em} Detection of HSV-1 antigen. An impression cytology smear obtained from a patient with HSK showing the presence of rounded up corneal epithelial cells positive for viral antigen. Infected cells show brilliant apple green fluorescence. Note the absence of background staining. Indirect immunofluorescence assay, × 500.  \\

\includegraphics[width=2.3cm, height=2.3cm]{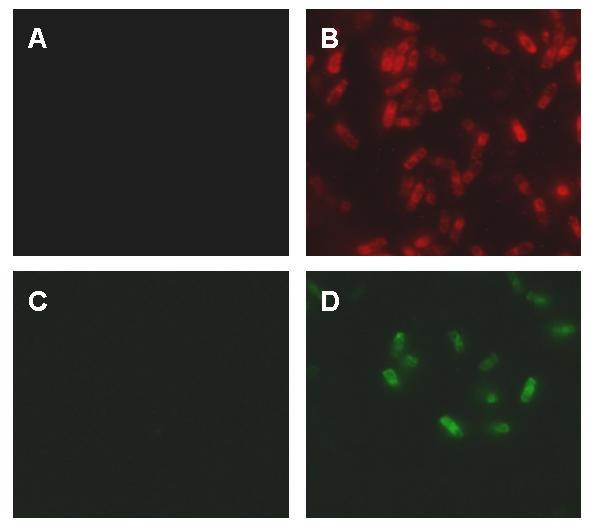} &    \vspace{-7.5em}  Detection of presence of CotC-LTB and CotC-TTFC by immunofluorescence microscopy. Sporulation of B. subtilis strains was induced by the resuspension method, and samples were taken 6 h after the onset of sporulation and analysed by immunofluorescence microscopy as described previously [46]. Samples were labelled with mouse anti-LTB antibody followed by anti-mouse IgG-TRITC conjugate (red fluorescein, Panels A \& B), or rabbit anti-TTFC antibody followed by anti-rabbit IgG-FITC conjugate (green fluorescein, Panels C \& D). Panels A \& C, wild type spores; Panel B, isogenic spores expressing CotC-LTB); Panel D, isogenic spores expressing CotC-TTFC.
  \\

\includegraphics[width=2.3cm, height=2.3cm]{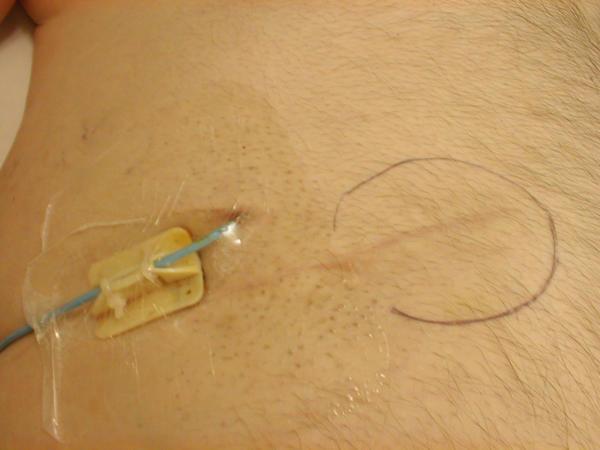} &  \vspace{-7em} Clinical photograph of the abdomen (close-up view): The surgical scar of implantation of baclofen pump is seen. The pigtail catheter emerges close to the scar. The skin around the pigtail catheter is red and angry-looking. Approximate position of baclofen pump is marked on the skin with a pen.\\
\includegraphics[width=2.3cm, height=2.3cm]{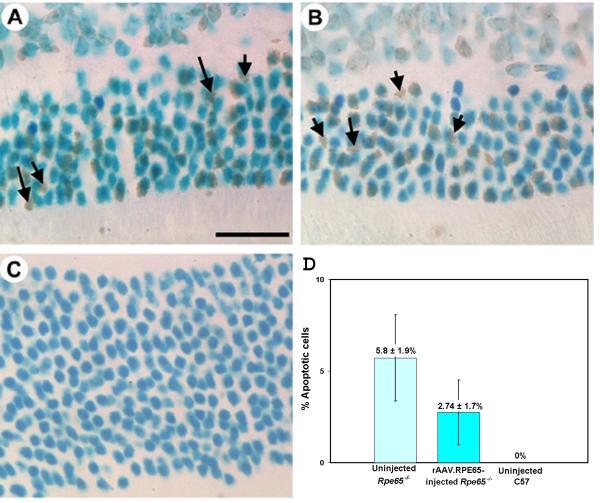}  & \vspace{-7.5em} Comparison of apoptotic cells numbers Photomicrographs of the outer nuclear layer of 8 mo uninjected Rpe65-/- (A), rAAV.RPE65 injected Rpe65-/- (B) and C57BL/6J mice (C) stained for apoptotic nuclei (arrows). (D) Graphical presentation of the percentage of photoreceptor nuclei that are apoptotic in uninjected Rpe65-/-, rAAV.RPE65-injected Rpe65-/- and uninjected C57BL/6J mice. Apoptotic and total photoreceptor nuclei were counted along 60 $\mu$m lengths of the outer nuclear layer of mice at 7 mo post-injection (8 mo of age). Average total photoreceptor counts: uninjected Rpe65-/- = 106.8 ± 22.9, rAAV.RPE65 injected Rpe65-/- = 134 ± 30.3, uninjected C57 = 213.5 ± 3.3. All data are mean ± S.D. 
\\

\includegraphics[width=2.3cm, height=2.3cm]{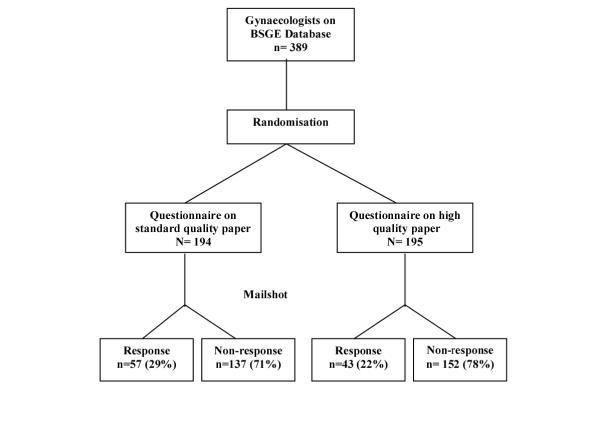}  & \vspace{-6.5em}
 Flow diagram showing randomisation and response rates of the survey.\\
 
\includegraphics[width=2.3cm, height=2.3cm]{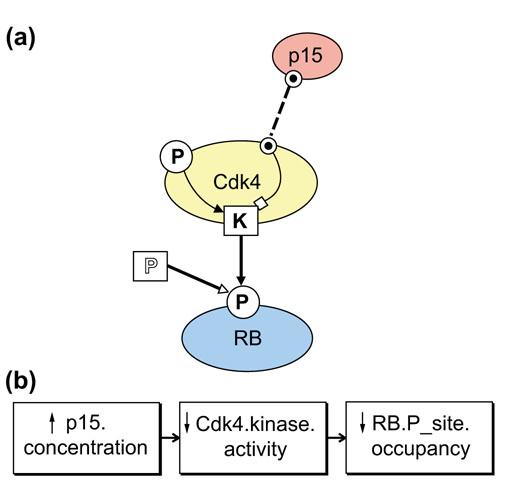}  &  \vspace{-7.5em} (a) A simplified model of a Cdk4 kinase molecule illustrates how basic BioD icons and action arrows can concisely represent intra- and inter-molecular actions. The Cdk4 molecule includes a kinase site (K) that, when active, phosphorylates a phosphorylation site (P) on the RB protein. The kinase site on Cdk4 is activated (filled arrow) by occupancy of its phosphorylation site and inhibited (open-squared arrow) by occupancy of the binding site (dotted circle) that binds the Cdk4 inhibitor p15. (b) An 'event model' derived from the model above. Events are defined as changes of state of one or more functional properties of icons in a state model. Here, for instance, the event model displays a chain of events triggered by an increase of p15 concentration (see text).\\

\includegraphics[width=2.3cm, height=2.3cm]{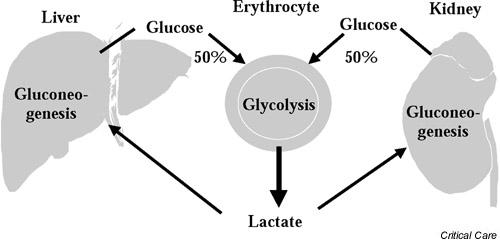}  & \vspace{-6.5em}
Renal contribution to endogenous glucose release from lactate during the postabsorptive phase. Data from [1].\\

\includegraphics[width=2.3cm, height=2.3cm]{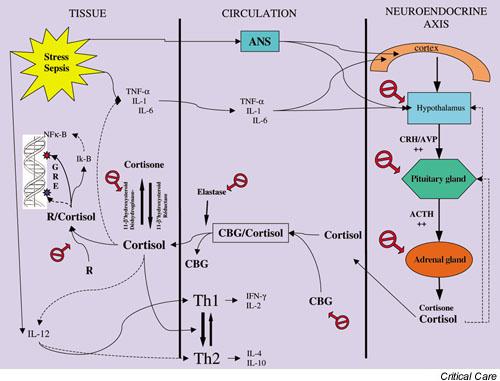}  & \vspace{-6.5em}
Strategy for detection and treatment of adrenal failure during sepsis. ACTH, adrenocorticotrophic hormone.\\

\includegraphics[width=2.3cm, height=2.3cm]{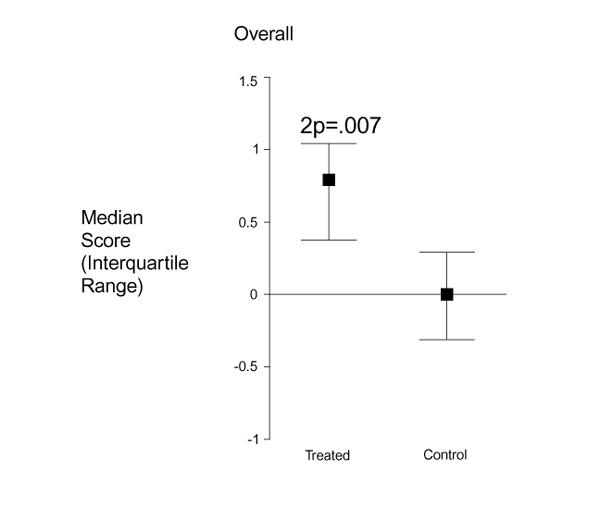}  & \vspace{-6.5em}
Mammographic changes in treated and control individuals. Values				are expressed as median change from baseline and interquartile range.\\

\bottomrule
\end{tabular}
\end{adjustbox}
\caption{Bomedica dataset Image-caption examples Benchmark Example: 10  examples from Biomedica dataset grouped by concept}
\label{biomedica-example}
\end{table*}

\begin{table*}[t]
\centering
\small
\begin{adjustbox}{max width=\textwidth}
\begin{tabular}{p{2cm}p{15.5cm}}
\toprule
\rowcolor{gray!10} \textbf{Image} & \textbf{Caption} \\
\includegraphics[width=2cm, height=2cm]{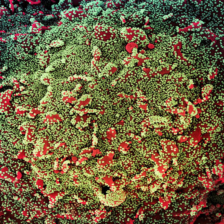}  &  \vspace{-5.5em} Colorized scanning electron micrograph of a cell (red) heavily infected with SARS-CoV-2 virus particles (green), isolated from a patient sample.  \\

\includegraphics[width=2cm, height=2cm]{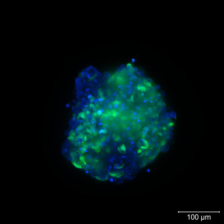}  &   \vspace{-5.5em}  HBE human derived cell lines cultured as micro-tissues (DAPI staining in blue; fixed with PFA). Infection with human Adenovirus type 2 expressing GFP.  \\

\includegraphics[width=2cm, height=2cm]{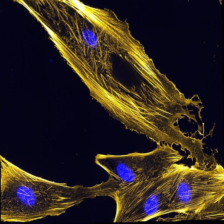} &    \vspace{-5.5em}  Immunofluorescence image of actin bundles in muscle precursor cells called myoblasts. The actin is labeled with fluorescently-tagged phalloidin, which is a toxin from the Amanita phalloides mushroom. Nuclei are shown in blue. 
  \\

\includegraphics[width=2cm, height=2cm]{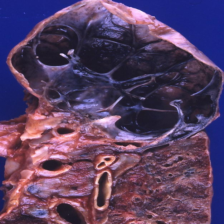} &  \vspace{-6em} Cut surface of a large apical bulla. Involved hilar lymph nodes also present.\\
\includegraphics[width=2cm, height=2cm]{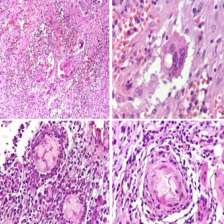}  & \vspace{-6em} In this pleural biopsy, a chronic inflammatory reaction with giant cells is seen, reacting to the presence of food material originating from an esophageal fistula. In the two bottom photographs, the structure of this material, although deteriorated, is still preserved. These are particles of vegetable material which, due to their size and shape, surely are seed -derived storage cells. \\
\includegraphics[width=2cm, height=2cm]{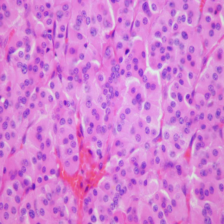}   & \vspace{-6em} Typical carcinoid tumor with organoid/insular growth and oncocytic tumor cells. There are many morphologic variants of carcinoid tumors. \\
\includegraphics[width=2cm, height=2cm]{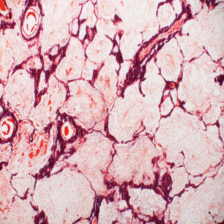}  & \vspace{-6em} Metastatic calcifiication of alveolar walls and blood vessels in an area of acute pneumonitis. \\
\includegraphics[width=2cm, height=2cm]{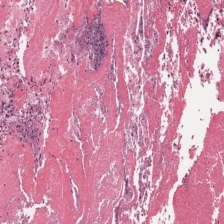}  & \vspace{-6em} Dicrofilarium \\
\includegraphics[width=2cm, height=2cm]{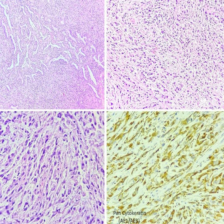} & \vspace{-7em}  Pleomorphic lung carcinoma is composed of 10\% or more of spindle cells and/or giant cells admixed with variable amounts of adenocarcinoma, squamous carcinoma, or large cell carcinoma. Some are composed solely of spindle cells and/or tumor giant cells. The diagnosis can only be made in a surgical specimen, not in a biopsy specimen. The type(s) of non-spindle cell carcinoma that are present should be mentioned in the pathology report. The term “sarcomatoid carcinoma” should be avoided because it is an umbrella term encompassing pleomorphic carcinoma, carcinosarcoma, and pulmonary blastoma.These images of pleomorphic carcinoma show malignant spindle cells admixed with adenocarcinoma \\

 \includegraphics[width=2cm, height=2cm]{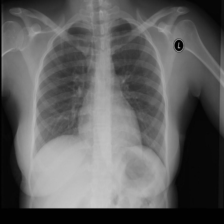}  & \vspace{-5em}  Normal PA chest x-ray 
  \\

\bottomrule
\end{tabular}
\end{adjustbox}
\caption{BiomedFlickr  Benchmark Example: 10 Random examples from biomedflcikr}
\label{flickr-example}
\end{table*}

\begin{figure*}[htb]
\small{

\begin{verbatim}
TAXONOMY = {'Ambiguous': ['ambiguous'],

    'Chemical Structures': ['2D chemical reaction','3D chemical reaction','2D chemical structure',
                            '3D protein structure','3D chemical structure'],

    'Clinical Imaging': ['x-ray radiography','optical coherence tomography','endoscopy',
                         'intraoral imaging','angiography','procedural image','skull','patient photo',
                         'functional magnetic resonance', 'magnetic resonance','eye','mammography',
                         'electrocardiography', 'clinical imaging', 'skin lesion','ultrasound',
                         'specimen','computerized tomography','laryngoscopy','teeth',
                         'intraoperative image','surgical procedure','brain'],

        
    'Graphs and Networks': ['graph','neural network','network'],

    'Illustrative Diagrams': ['sankey diagram','metabolic pathway','scientific illustration','diagram',
                              'signaling pathway','illustrative diagram','flow diagram',
                              'cohort selection flowchart','illustration','drawing','system diagram',
                              'flowchart'],

    'Immuno Assays': ['immunocytochemistry','karyotype','gel electrophoresis','immunoassay',
                      'immunoblot','assay','immunohistochemistry'],

    'Laboratory Specimens and Cultures': ['reagents','laboratory specimen','bacterial culture'],

    'Maps': ['map'],

    'Microscopy': ['scanning electron microscopy','electron microscopy','flowcytometry',
                   'transmission electron microscopy','light microscopy','fluorescence microscopy',
                   'phase contrast microscopy','confocal microscopy','epifluorescence microscopy',
                   'microscopy'],

    'Natural Images': ['face','aerial photography','natural image','human head','humans and devices',
                       'human','insects','nature'],
    
    'PCR': ['qPCR','RT PCR'],


    'Plots and Charts': ['violin plot','bar plot','roc curve','sequence plot','radial plot','plot',
                         'matrix plot','phylogenetic tree','process chart','dot plot','pyramid chart',
                         'forest plot','box plot','survival curve','circos plot','venn diagram',
                         'heatmap plot','circular plot','scatter plot','word cloud','list','tree',
                         'density plot','funnel plot','plot and chart','2D mesh','3D plot',
                         'radial diagram','pie chart','manuscript','histogram',
                         'differential gene expression matrix','line plot','signal plot'],

        
    'Screen Based Visuals': ['screenshot','user interface'],
    
    'Scientific Formulae and Equations': ['algorithm'],
    
    'Tables': ['table','checklist table'],
    
    'Tools and Materials': ['medical equipment','microscope','electronic circuit','lab equipment',
                            'tool']}
\end{verbatim}}
\caption{Hierarchical Taxonomy. Filtered hierarchical taxonomy with topics included in the BIOMEDICA dataset.}

\label{fig:xml_markup_example}
\end{figure*}

\end{document}